\documentclass[letterpaper, 10 pt, journal, twoside]{IEEEtran} 
\usepackage{caption}
\usepackage[utf8]{inputenc}
\usepackage[T1]{fontenc}

\usepackage{graphics} 
\usepackage{epsfig} 
\usepackage{bm}
\usepackage{times} 
\usepackage{amsmath} 
\usepackage{amssymb}  
\usepackage{float}
\usepackage[space, compress]{cite}
\usepackage{mathtools}
\usepackage{graphicx}
\usepackage{hyperref}
\usepackage[font=footnotesize,labelfont=bf]{caption}
\usepackage{subcaption}
\usepackage{makecell}
\usepackage{upgreek}
\usepackage[dvipsnames]{xcolor}
\usepackage{pbox}
\usepackage{soul}

\usepackage{enumitem}
\usepackage[version=4]{mhchem}
\usepackage{multirow}
\usepackage{setspace, lipsum}
\usepackage{babel}
\usepackage{blindtext}
\usepackage{booktabs}
\usepackage{threeparttable}
\usepackage{tabularx}

\usepackage{algorithm} 
\usepackage{algpseudocode} 

\newcounter{qsubsection}

\newcommand{\qsubsection}[2]{
\refstepcounter{qsubsection}
\setcounter{subsubsection}{0}
\subsection*{Q{#1}.\quad #2}
}%

\author{Po-Yen Wu, Cheng-Yu Kuo, Yuki Kadokawa, and Takamitsu Matsubara%
\thanks{All author are with Division of Information Science, Graduate School of Science and Technology, Nara Institute of Science and Technology, Nara 630-0192, Japan}
\thanks{This work is supported by JST Moonshot Research and Development, Grant Number JPMJMS2032}
}

\title{Prolonging Tool Life:\\ Learning Skillful Use of General-purpose Tools through Lifespan-guided Reinforcement Learning}
\newcommand*{\fref}[1]{Fig.~\ref{#1}}  
\newcommand*{\eref}[1]{(\ref{#1})}            

\begin{document}
\maketitle
\begin{abstract}
In inaccessible environments with uncertain task demands, robots often rely on general-purpose tools that lack predefined usage strategies. These tools are not tailored for particular operations, making their longevity highly sensitive to how they are used. This creates a fundamental challenge: how can a robot learn a tool-use policy that both completes the task and prolongs the tool’s lifespan? In this work, we address this challenge by introducing a reinforcement learning (RL) framework that incorporates tool lifespan as a factor during policy optimization. Our framework leverages Finite Element Analysis (FEA) and Miner’s Rule to estimate Remaining Useful Life (RUL) based on accumulated stress, and integrates the RUL into the RL reward to guide policy learning toward lifespan-guided behavior. To handle the fact that RUL can only be estimated after task execution, we introduce an Adaptive Reward Normalization (ARN) mechanism that dynamically adjusts reward scaling based on estimated RULs, ensuring stable learning signals. We validate our method across simulated and real-world tool use tasks, including Object-Moving and Door-Opening with multiple general-purpose tools. The learned policies consistently prolong tool lifespan (up to 8.01$\times$ in simulation) and transfer effectively to real-world settings, demonstrating the practical value of learning lifespan-guided tool use strategies.
\end{abstract}
\begin{figure}[t]
    \centering
    \includegraphics[width=1.0\linewidth]{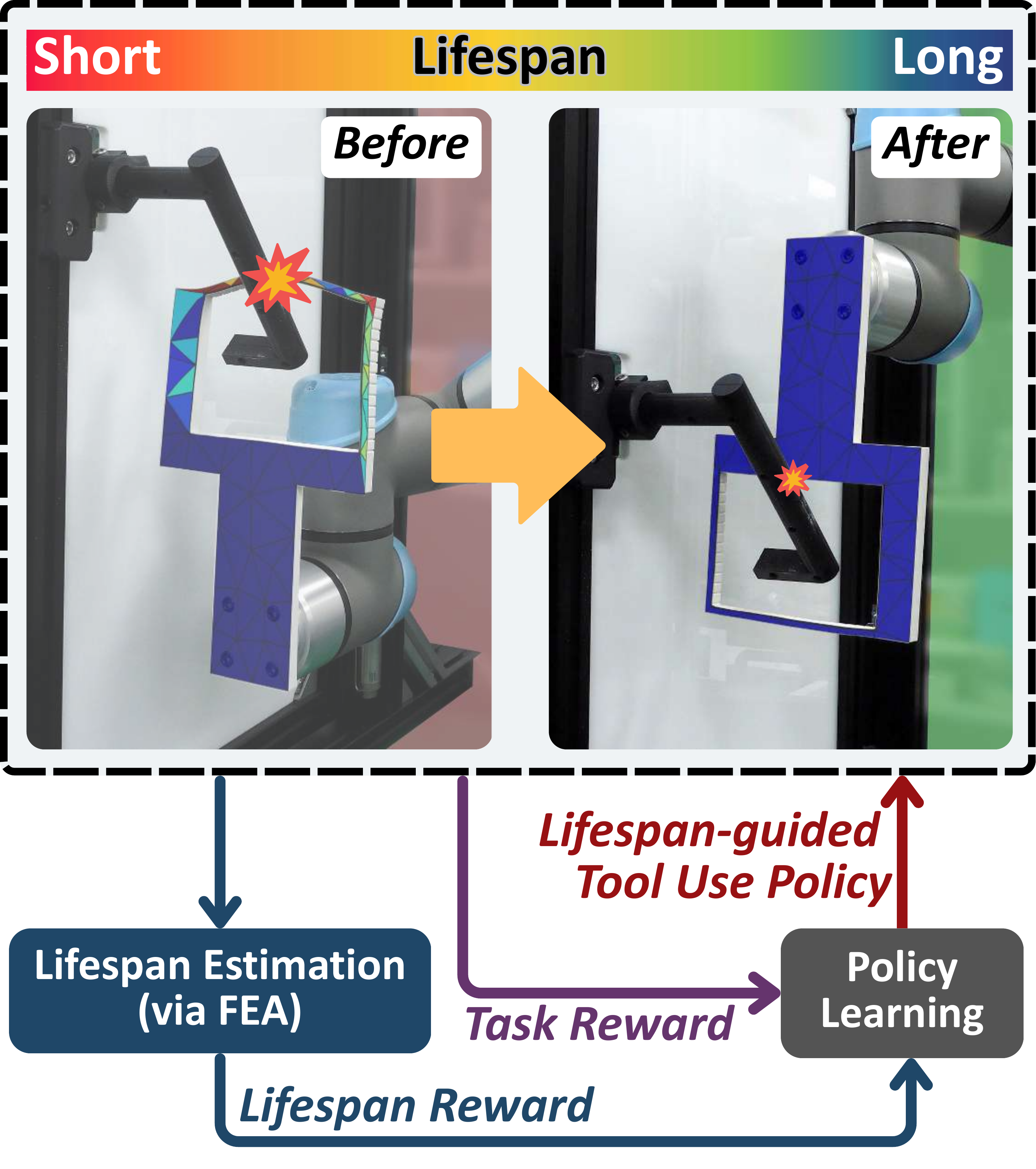}
    \caption{\textbf{An illustration of how incorporating tool lifespan into policy learning enables the development of lifespan-guided tool-use strategies.} When lifespan is not considered (Before), the learned policy does not account for structural variations across the tool, often applying stress to weaker regions and causing early failure. By integrating lifespan estimation (via FEA) into reinforcement learning, the agent receives a life reward that guides it toward structurally robust regions (After). The resulting lifespan-guided tool use policy balances task reward and lifespan reward to achieve both task success and extended tool lifespan.}
    \label{fig:Intro}
\end{figure}

\section{Introduction}
\label{sec:introduction}
\IEEEPARstart{R}{obots} are increasingly deployed in inaccessible scenarios, such as lunar surfaces \cite{Chen2024lunar}, ruins \cite{zhang2022ruin}, and mining sites \cite{Fernando2020Mining}. These environments are characterized by environmental uncertainties, which arise a wide range of potential tasks and their conditions. It is impossible to equip robots with task-specific tools for each individual scenario. Instead, robots must rely on general-purpose tools, which are not tailored for specific operations and don't have a preferred usage to complete the task. However, non-optimal usage conditions increase the failure rate of these tools, and replacing damaged components is both time-consuming and costly, significantly reducing operational efficiency and effectiveness~\cite{YOU2025Construction}. Therefore, in inaccessible scenarios involving general-purpose tool use, optimizing tool use policies to enhance tool lifespan can substantially improve operational efficiency. This motivates a critical research question: How can a robot learn a tool use policy to maximize the tool's lifespan while completing the task?

Learning robotic tool use skill is a process of acquiring task-completing policy according to various environmental conditions, such as the tool's geometry, environmental constraints, and task requirements. By introducing an additional environmental factor, we can guide the tool use policy to improve performance metrics such as motion stability~\cite{holladay2019force, qin2020keto} and task efficiency~\cite{zhang2022understanding}. In this study, we adopt a similar approach of leveraging environmental factors to incorporate tool lifespan as a key performance metric alongside task completion, explicitly considering damages (such as material fatigue and structural wear) during the process of tool use policy optimization. In practice, however, combining learning methods, such as Reinforcement Learning (RL)~\cite{kaelbling1996reinforce}, with the tool's lifespan raises several challenges: (1) quantifying the tool's lifespan from loads; (2) integrating lifespan considerations into the learning of task-completing tool-use policies using RL; and (3) addressing the classic ``chicken-and-egg'' challenge in reward engineering, where lifespan estimates needed for reward design are only obtainable after execution.

To address the above challenges, we propose a RL framework with the following elements: (1) Leveraging Finite Element Analysis (FEA) with Miner's rule to accumulate cyclic load for quantifying potential damage and estimating the tool's Remain Useful Life (RUL) for evaluating policy's quality. (2) Incorporating RUL into the reward function to optimize policies that ensure both successful task completion and lifespan extension. (3) We introduce an Adaptive Reward Normalization (ARN) mechanism to resolve the issue that RUL is only estimated after task execution, which makes static reward normalization unreliable. By adjusting the normalizing rule based on observed RULs from previous episodes, ARN ensures stable reward signals throughout learning. Together, these components enable the learning of a tool use policy that goes beyond task success, incorporating lifespan-guided strategies that extend the tool's lifespan, as illustrated in Fig.~\ref{fig:Intro}.

We validate our framework in both simulation and real-world environments, demonstrating its ability to learn tool use policies that extend tool lifespan across Object-Moving and Door-Opening tasks with multiple general-purpose tools. As FEA and RUL estimation are only feasible in simulation, we first constructed simulated environments aligned with physical setups. Object-moving policies were trained using four tool geometries, while Door-Opening policies used two of them. Compared to task-only baselines, our lifespan-guided policies consistently achieved longer tool lifespan (up to 12.54$\times$) in simulation. We then deployed these policies on physical robots via sim-to-real transfer, repeatedly executing tasks until tool failure. The real-world results confirm that our method effectively prolongs tool lifespan while retaining task completion capability, validating its practicality for robotic tool use applications. Building on these findings, our key contributions are as follows:
\begin{enumerate}
    \item A lifespan-guided RL framework integrating FEA-based RUL to balance task success and tool longevity.
    \item An adaptive reward normalization mechanism supporting robust learning without preset lifespan bounds.
    \item  Real-world validation using physical tool failure counts, showing the method’s practical value.
\end{enumerate}
\section{Related work}
\subsection{Applications of Guiding Robotic Tool Use Policies}
The diversity of robotic tool use (involving various tasks, tool types, and environmental contexts) has led to a wide range of research directions. Some studies leverage language models to provide robots with semantic guidance about tool geometry, thereby accelerating the learning of tool use policies~\cite{ren2023leveraging,xu2023creative}. Others investigate how to model tool-environment interactions to improve the robot’s ability to select appropriate tools~\cite{saito2021select}.

While these approaches emphasize learning tool use policies under diverse conditions, another line of work focuses on shaping tool use behaviors to achieve specific performance characteristics. For example, researchers have incorporated kinematic constraints to enhance manipulation stability~\cite{holladay2019force}, applied object-tool keypoint alignment to improve precision~\cite{qin2020keto, Turpin2021GIFT}, and modeled physical effects to boost operational efficiency~\cite{zhang2022understanding}. 

In contrast, our work incorporates damage estimation into the learning loop, aiming to extend tool lifespan as a performance criterion in robotic tool use.

\subsection{FEA in Robotics}
Finite Element Analysis (FEA) is a widely used numerical method for simulating physical behaviors in robotics~\cite{elsayed2014finitefluid, zhang2022understanding}. It is commonly applied to (1) evaluate stress distributions for failure prediction and design optimization~\cite{ding2022dynamicsoft, xavier2022softreview, KOURITEM202212847}, (2) simulate deformation to support stable control during manipulation~\cite{duriez2013controlsoft, ficuciello2018femcontrol}, and (3) model interactions with deformable objects by estimating internal physical states for control evaluation~\cite{huang2022defgraspsim, mykhailyshyn2023finitegrasp}. 

While prior studies focus on design or control applications, we integrate FEA-based damage estimation into reinforcement learning, using it as a reward signal to promote tool use policies that reduce wear and extend lifespan.

\begin{figure*}[ht]
    \centering
    \includegraphics[width=1.0\hsize]{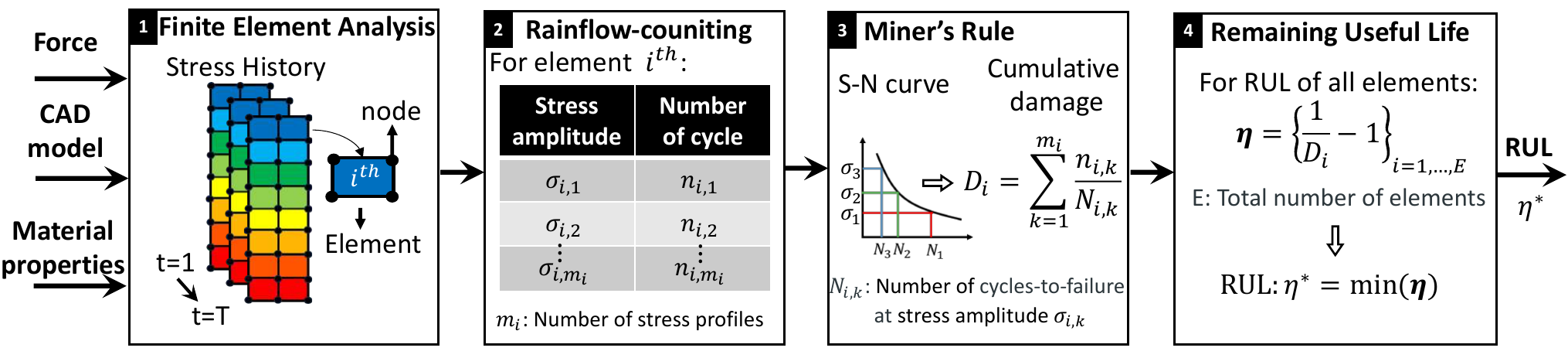}
    \caption{\textbf{Remaining Useful Life calculation flow.} RUL calculation can be divided into 4 steps: (1) Finite Element Analysis (FEA) to generate a stress history based on the tool's CAD model, material properties, and force data. (2) Rainflow-counting to identify stress amplitudes and their respective cycles from the stress history. (3) S-N curves determine the number of cycles to failure for each stress amplitude, allowing Miner's rule to evaluate the cumulative damage of the elements on the tool. (4) Derive the tool's RUL based on the inverse of maximum cumulative damage.}
    \label{fig:RUL_cal}
\end{figure*}

\subsection{Lifespan-guided Control in Automation}
Lifespan-guided control has been studied in various domains to extend system longevity while maintaining functional performance. In wind energy systems, individual pitch control reduces fatigue damage in turbine blades under high wind conditions~\cite{coquelet2022wind}. In cutting processes, strategies such as dynamic spindle speed control~\cite{VAVRUSKA2023283} and cutting parameter optimization~\cite{CHOUDHURY1999343} are employed to minimize cutting-tool wear. In actuator systems, reinforcement learning has been used to optimize motor operating longevity~\cite{JHA2019HAC}. Similarly, robotic arms have adopted load-balancing strategies to reduce joint fatigue~\cite{peternel2018fatiguemanage, costa2022prolonging}. 

Differently, our work focuses on extending the lifespan of external tools in robotic manipulation by optimizing their usage strategies.

\section{Preliminaries}
\subsection{Reinforcement Learning}
We aim to learn a tool use policy that balances task completion and tool lifespan. The complexity of contact dynamics and non-linear material damage makes it difficult to model this problem analytically. Instead, we adopt RL, which enables data-driven policy optimization through interaction in such environments.

In this work, the RL problem is formulated as a Markov Decision Process, defined by the tuple \(\mathcal{M} = (\mathcal{S}, \mathcal{A}, \mathcal{T}, \mathcal{R}, \gamma)\)~\cite{kaelbling1996reinforce}. Here, \(\mathcal{S}\) is the set of possible environment states, and \(\mathcal{A}\) is the set of actions available to the agent. \(\mathcal{T}_{ss'}^a\) is the probability of transition from state \(s\) to state \(s'\) after taking action \(a\), the corresponding reward of making the transition is represented as \(\mathcal{R} = r_{ss'}^a\), and \(\gamma \in [0,1)\) is the discount factor. Policy $\pi(a|s):\mathcal{S} \rightarrow \mathcal{P}(\mathcal{A})$ is the probability of choosing action $a$ given state $s$. The \emph{Q-value} \(Q_\pi: \mathcal{S} \times \mathcal{A} \rightarrow \mathbb{R}\) is defined as the expected total discounted reward starting from state $s$, taking action $a$, and thereafter following policy $\pi$:
\begin{align}
\label{eq:q-function}
Q_\pi(s, a) = \mathbb{E}_{\pi, \mathcal{T}} \left[ \sum_{t=0}^{\infty} \gamma^t r_{s_t} \,|\, s_0 = s, a_0 = a \right]
\end{align}
where \(r_{s_t} = \sum_{\substack{a_{t} \in \mathcal{A} \\ s_{t+1} \in \mathcal{S}}} \pi(a|s_t)\mathcal{T}_{s_{t}s_{t+1}'}^a r_{s_ts_{t+1}'}^a\) is the reward at state \(s\). The goal of RL is to find an optimal policy \(\pi^*\) that satisfies the Bellman equation:
\begin{align}
\label{Qfunc}
Q^{*}(s,a) = \max_\pi \sum_{s' \in \mathcal{S}}\mathcal{T}_{ss'}^{a}(r_{ss'}^a+\gamma \sum_{a' \in \mathcal{A}} \pi(a'|s')Q^*(s',a')),
\end{align}
where $Q^*$ is an optimal $Q$ function.

In this work, we employ Soft Actor-Critic (SAC)~\cite{haarnoja2018}, an off-policy reinforcement learning algorithm that optimizes a policy \(\pi_{\theta}\) to maximize the expected return while incorporating policy entropy to enhance exploration. In SAC, the soft Q-function is defined as:
\begin{align}
\label{eq:SAC_Q}
Q_{\pi_{\theta}}(s,a) &= \mathbb{E}_{\substack{s' \sim \mathcal{T}_{ss'}^a \\ a' \sim {\pi_{\theta}}}} \Big[ r_{ss'}^a + \gamma (Q_{\pi_{\theta}}(s', a') - \alpha \log {\pi_{\theta}}(a'|s') \Big],
\end{align}
where \(\alpha > 0\) is a temperature parameter controlling the trade-off between expected return and policy entropy. The policy is updated by minimizing the following loss function:
\begin{align}
\label{eq:SAC_loss}
L_\pi(\theta) = \mathbb{E}_{a \sim \pi_{\theta}(\cdot|s)} \left[ \alpha \log \pi_{\theta}(a|s) - Q_{\pi_{\theta}}(s, a) \right].
\end{align}
SAC iteratively updates the policy parameters \(\theta\) and the Q-function to minimize this entropy-regularized objective, often tuning \(\alpha\) adaptively to balance exploration and exploitation.

\subsection{Lifespan Estimation}
To enable a learning agent to optimize for tool longevity, a quantitative measure of the tool's lifespan must be provided as a feedback signal. However, unlike properties such as external forces, a tool's lifespan cannot be directly measured by sensors and must instead be estimated. Therefore, we adopt a standard procedure from damage analysis ~\cite{chung2016FRM1,baek2008FRM2} to acquire this metric:  (1) Using Finite Element Analysis (FEA) ~\cite{szabo2021finite} to determine stress distributions, (2) applying the rainflow-counting algorithm ~\cite{matsuishi1968rain} to parse stress cycles, and (3) utilizing Miner's Rule ~\cite{miner1945} to calculate cumulative damage and quantify the tool's lifespan as the Remaining Useful Life (RUL). The RUL calculation flow is shown as \fref{fig:RUL_cal}.

\subsubsection{Finite Element Analysis (FEA)}
To evaluate the tool's internal forces, FEA estimates internal stress distributions within tool geometries by discretizing the structure into a mesh of finite elements. Leveraging the tool's CAD model, material properties, and external force history data, FEA solves the governing equilibrium equations to obtain the stress history at each element over the simulation period $t \in [0, T]$ where $T$ represents the total simulation time and $\sigma_i(t)$ being the time-varying stress at element $i$, denoted as $\left\{\sigma_i(t)\right\}_{i=1..E}$, where $E$ is the total number of elements in the tool.

\subsubsection{Rainflow Counting}
Using a tool to perform tasks can result in complex and non-periodic stress histories. This poses a challenge for damage evaluation, as damage quantification relies on evaluating discrete load cycles. The rainflow-counting algorithm resolves this by decomposing complex stress sequences into a rainflow series (pairs of equivalent loads and corresponding cycle count). Applying rainflow counting to the FEA-provided stress history $\sigma_i(t)$ at each element $i$ yields a set of cycle counts associated with distinct stress profiles (amplitude):
\begin{align}
    \Bigl\{(\Delta \sigma_{i,k},\, n_{i,k})\Bigr\}_{k=1,..,m_{i}}
\end{align}
where $m_i$ is the number of unique stress profiles levels observed at $i^{th}$ element, $\Delta \sigma_{ik}$ is the $k$th amplitude, and $n_{ik}$ denotes the corresponding number of cycles.

\begin{figure*}[t]
    \centering
    \includegraphics[width=1\hsize]{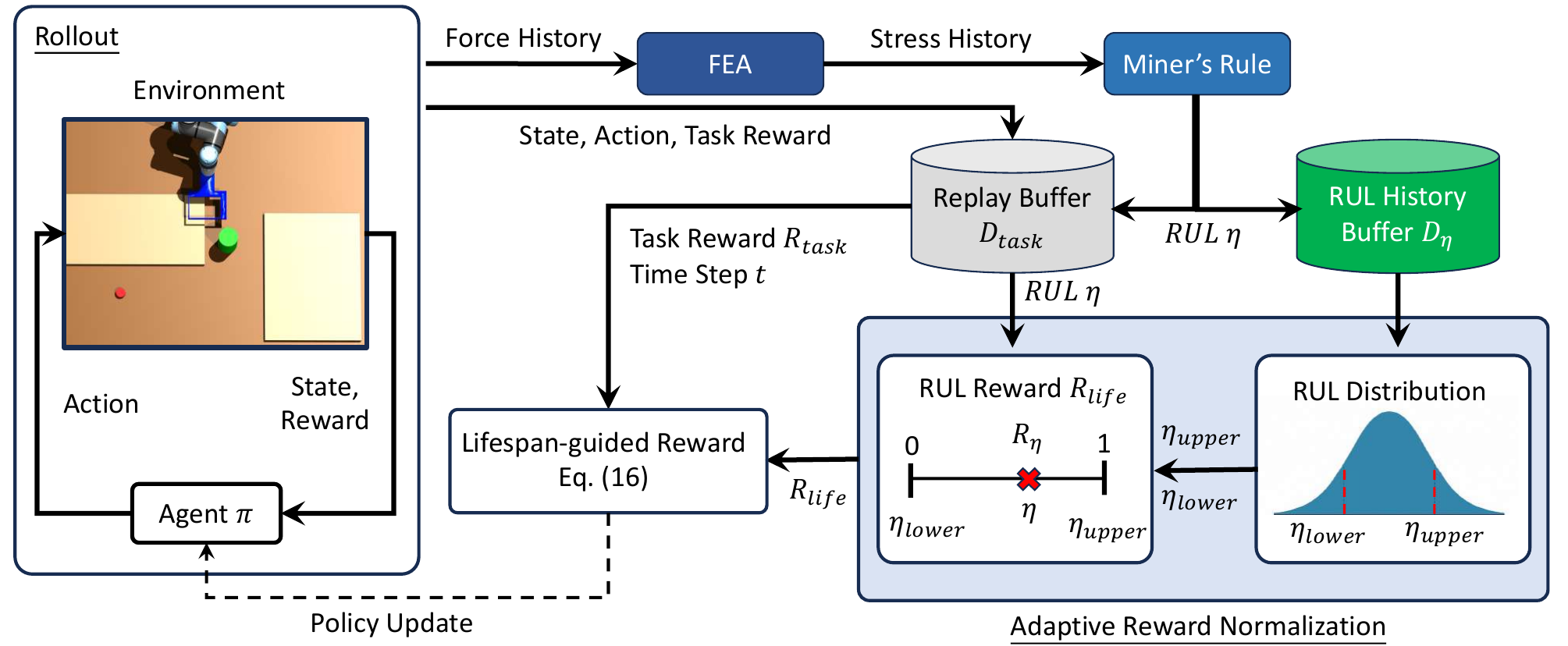}
    \caption{
    \textbf{Overview of the proposed method integrating lifespan-guided reward into reinforcement learning.} During each rollout, the agent interacts with the environment, collecting state, action, and task reward information. At the end of each episode, the stress history is calculated via finite element analysis (FEA) and processed with Miner’s rule to estimate the remaining useful life (RUL). The RUL value is stored in a history buffer, which is used by the adaptive reward normalization (ARN) mechanism to determine dynamic upper and lower bounds. These bounds are applied to normalize the life reward for subsequent episodes, ensuring stable and meaningful reward signals for policy learning.
    }
    \label{fig:Proposed_method}
\end{figure*}

\subsubsection{Cumulative Damage via Miner's Rule}
Given the rainflow series, the accumulated damage $\boldsymbol{D}=\left\{D_{i}\right\}_{i=1,..,E}$ (a series of elemental damage) is derived by Miner's rule, which linearly aggregates each element's damage from all its load cycles:
\begin{align}
\label{Miner's rule}
    D_{i}
    \;=\; 
    \sum_{k=1}^{m_i}\;\frac{n_{ik}}{N_{\Delta \sigma_{ik}}},
\end{align} 
where $D_i\in\mathbb{R}_+$ is the estimated damage at element $i$, while $D_{i}\;\ge\;1$ indicates that the element has theoretically been damaged. $N_{\Delta \sigma_{ik}}$ is the corresponding number of cycles-to-failure, obtained via Basquin’s law~\cite{oh1910BASQUIN}:
\begin{align}
\label{Basquin}
N_{\Delta \sigma_{i,k}} \;=\; \mathrm{a}\,\bigl(\Delta \sigma_{i,k}\bigr)^{-\mathrm{b}},
\end{align}
where $\mathrm{a}$ and $\mathrm{b}$ are material-specific constants derived from its S-N curve.

\subsubsection{Estimation of Remaining Useful Life}
Given the series of elemental damage $\boldsymbol{D}$, we can evaluate all element's Remaining Useful Life (RUL, denoted as $\boldsymbol{\eta}$) by counting the expected load cycles it can sustain under the given stress condition:
\begin{align}
    \begin{split}
    \label{eq:RUL_element}
        \boldsymbol{\eta} &= \left\{\frac{1}{{D_i}}-1
        \right\}_{i=1,..,E}
    \end{split}.
\end{align}
Given that the failure of any single element may lead to tool failure, we define the tool's RUL $\eta^*$ as:
\begin{align}
    \begin{split}
    \label{eq:RUL}
        \eta^* &= \text{min}\left(\boldsymbol{\eta}\right)
    \end{split}.
\end{align}

\section{Proposed Method}\label{sec:Proposed_method}
\subsection{Overview}
In this section, we introduce the proposed lifespan-guided tool use reinforcement learning framework, designed to optimize tool use policies for task completion while extending general-purpose tool's lifespan. The approach integrates RL with mechanical analysis feedback, where tool degradation is quantified as Remaining Useful Life (RUL) estimated via FEA~\cite{szabo2021finite} and Miner’s Rule~\cite{miner1945}. This RUL estimate is then used to construct an \emph{RUL Reward} that explicitly encourages lifespan extension.

In parallel, we define a \emph{Task Reward} to evaluate successful task completion. These two components are combined to form a \emph{Lifespan-guided Reward}, which guides the agent to complete tasks effectively while minimizing structural damage and prolonging the tool's usable life.

To address the challenge of unpredictable lifespan variations across different tools and tasks, we introduce Adaptive Reward Normalization (ARN). ARN adaptively scales reward signals based on the history of observed RUL from successful episodes and updates all previous rewards accordingly before each training step. This mechanism ensures stable learning signals during policy optimization and accommodates variability in RUL across different setups. The overall architecture of the framework is illustrated in \fref{fig:Proposed_method}.

\subsection{Lifespan-guided Reward Design}
\subsubsection{Reward Function Formulation}
To encourage the learning of policies that achieve task completion while extending RUL, we define a \emph{Lifespan-guided Reward} function over state-action-RUL tuples ($R(s,a, \eta):\mathcal{S}\times\mathcal{A}\times\mathbb{R}\rightarrow\mathbb{R}$) as:
\begin{align}
    \begin{split}                           
    \label{eq:TRUL}
        R(s,a,\eta) := 
        \begin{cases}
        R_{task}(s,a) & \text{if} \ t <T\\
        R_{task}(s,a)+ R_{life}(\eta) &  \text{if} \ t = T 
        \end{cases}
    \end{split} \ \ ,
\end{align}
where $t \in [0, T]$, and $T$ is the episode length. $R_{task}(s,a):\mathcal{S}\times\mathcal{A}\rightarrow\mathbb{R}$ is the \emph{Task Reward} that evaluates task completion at each step, while $R_{life}(\eta): \mathbb{R} \to \mathbb{R}$ represents the \emph{RUL Reward} provided only at episode termination, as it depends on the complete stress history for cumulative damage assessment.

We define $R_{life}(\eta)$ as a normalized reward that reflects the achieved tool lifespan:
\begin{align}
    \label{eq:RUL_norm}
        R_{life}(\eta) := \mathbb{I}\{\text{task completed}\} \cdot
        \frac{\eta-\eta_{lower}}{\eta_{upper}-\eta_{lower}},
\end{align}
where $\boldsymbol{\eta}_b:=\left\{\eta_{upper}, \eta_{lower}\right\} \in \mathbb{R}$ denotes the normalization bounds. $\mathbb{I}\{\text{task completed}\}$ is an indicator function equal to $1$ for successful episodes and $0$ otherwise. This ensures that only data from completed tasks contributes to policy updates, preventing misleading gradients from failed episodes with high RUL.

However, direct implementation of the \emph{RUL Reward} can pose several issues:
\begin{enumerate}
    \item As defined in \eref{eq:TRUL} and \eref{eq:RUL_norm}, the \emph{RUL Reward} is only available at episode termination and is further gated by task success, leading to significant reward sparsity. This infrequent feedback severely limits learning efficiency and the policy's ability to discover strategies that improve lifespan, motivating the need for mechanisms that can densify and effectively propagate lifespan-related signals.
    
    \item Determining appropriate $\eta_{upper}$ and $\eta_{lower}$ for normalization remains challenging. The true RUL range depends on the specific tool and task and is only observable after episodes. Relying on static bounds risks suboptimal scaling or near-zero gradients, motivating the need for adaptive normalization that adjusts bounds based on observed RUL outcomes.
    
    \item A key challenge in adaptive normalization is that the estimated reward boundaries can shift abruptly between episodes, especially early in training or when encountering outlier RUL values. Such sudden changes in normalization bounds can cause large shifts in the reward distribution, destabilizing policy learning and leading to poor convergence. This underscores the need for adaptive normalization with smoothing strategies to ensure gradual updates and maintain training stability.
\end{enumerate}

\subsubsection{Densifying Sparse Rewards}  

To address the reward sparsity described earlier, we introduce a reward redistribution technique that densifies the training signal, drawing inspiration from TD($\lambda$)'s~\cite{TD1998Sutton} approach to temporal credit assignment. While TD($\lambda$) propagates the impact of TD errors backward through eligibility traces to update value estimates, our method operates directly at the reward level by redistributing the final \emph{RUL Reward} across preceding steps. This strategy ensures that lifespan-related feedback is effectively utilized during learning, enabling the agent to associate early actions with their long-term impact on tool damage.

For each transition in the replay buffer, we record its timestep $t$ within the trajectory. During training, when sampling transitions, we retrieve the episode's terminal \emph{RUL Reward} and compute an effective training reward:
\begin{align}
\label{eq:credit_assign}
    R(s,a,\eta) := R_{task}(s,a) + \gamma_{b}^{T-t} R_{life}(\eta),
\end{align}
where $t \in [0, T]$ and $\gamma_{b} \in [0,1]$ are decay factors controlling the reward’s temporal spread. This approach reduces learning uncertainty, connects lifespan feedback to earlier decisions, and discourages suboptimal strategies that sacrifice tool lifespan for short-term task success.

\begin{algorithm}[t]
\caption{Lifespan-guided Tool Use Reinforcement Learning}
\label{alg:LATURL}
\begin{algorithmic}[1]
\State Initialize $\phi$, replay buffer $D$, lifespan buffer $D_{\eta}$.
\State Set parameters described in Table.~\ref{tab:params}
\For{episode $e$ = 0,...,$E$}
    \For{environment step $t$ = 0,...,$T$}
        \State Execute $a_t \sim \pi_{\phi}(a_t|s_t)$
        \State Observe next state $s_{t+1}$ and reward $r_t$
        \State Get transition $\tau_{t} = {(s_t,a_t,r_t,s_{t+1},t)}$
        \State $D \gets D \cup \tau_{t}$
        \For{each gradient step}
            \State $\{(s_t, a_t, r_t, s_{t+1}, t, \eta)\} \sim D$
            \State Calculate training reward $R^{(e)}_{ARN}$ by \eref{eq:DRFP}
            \State Update policy $\pi_{\phi}$ with ${(s_t, a_t, R^{(e)}_{ARN} , s_{t+1})}$
        \EndFor
    \EndFor
    \State Compute RUL $\eta$
    \State $D_{\eta} \gets D_{\eta} \cup \{\eta\}$
    \For {transition $\tau_{i}$ from episode $e$ in $D$}
        \State Update $\tau_i$ to {($s_i,a_i,r_i,s_{i+1},t_i,\eta$)}
    \EndFor
    \State Compute Boundaries $\boldsymbol{\eta}_{b}^{e}$ by \eref{eq:smooth_bound}
\EndFor
\end{algorithmic}
\end{algorithm}

\subsubsection{Adaptive Reward Normalization (ARN): Adaptive normalization of RUL reward}  
To handle the challenge of unknown and varying RUL ranges across different tools and tasks, we adopt an experience-based adaptive normalization strategy. After each successful episode, the observed RUL is stored in a buffer $D_{\eta}$ that maintains the history of recent episodes (set to 500 in this work). This rolling history allows the normalization bounds to evolve in line with the agent’s learning progress, providing appropriately scaled rewards throughout training.

To compute the normalization bounds, we use a percentile-based approach that improves robustness against outliers when estimating RUL:
\begin{equation}
\left.
    \begin{aligned}
    \hat{\eta}_{upper} &:= \text{Percentile}(D_{\eta}, \, 100-\beta) \\
    \hat{\eta}_{lower} &:= \text{Percentile}(D_{\eta}, \, \beta)
\end{aligned}\right\}.
\label{eq:percentile_bound}
\end{equation}
Here, $\beta$ is the percentile factor, set to 5 in this work. This method ensures that normalization bounds adapt to the distribution of recent RUL outcomes, reducing sensitivity to outlier episodes.

\begin{table}[t]
\caption{\textbf{Parameter Settings Used in the Experiments}}
\label{tab:params}
\setlength{\tabcolsep}{3pt}
\begin{tabular}{|p{35pt}|p{140pt}|p{55pt}|}
\hline
\textbf{Parameter} & \textbf{Description} & \textbf{Value} \\
\hline
$B$ & Minibatch size & 32 \\
$E$ & Number of episodes & 2500 \\
$T$ & Number of steps per episode & 20 \\
$T$ & Steps per episode (Door-Opening) & 6 \\
$\gamma$ & Discount factor of RL & 0.99 \\
$\beta$ & Percentile factor & 5 \\
$\alpha_s$ & Smooth update factor & 0.2 \\
$\gamma_b$ & Decay factor of credit assignment & 0.95 \\
$\gamma_b$ & Decay factor (Door-Opening) & 0.9 \\
\hline
\end{tabular}
\end{table}

\subsubsection{Smoothing Normalization Updates}  
Adaptive normalization can lead to abrupt changes in estimated reward boundaries between episodes, causing large variations in the reward distribution that destabilize policy learning. To address this, we apply exponential moving average smoothing after each episode to ensure gradual updates to the normalization bounds. This approach stabilizes reward scaling across episodes, reduces learning variance, and supports consistent policy convergence as the agent adapts to evolving RUL distributions.

After each episode, the smoothed boundaries for episode $e$, denoted $\boldsymbol{\eta}^{e}_b = \{\eta_{upper}^{e}, \eta_{lower}^{e}\}$, are updated as follows:
\begin{equation}
\boldsymbol{\eta}^{e}_b:=
\left\{
    \begin{aligned}
    \eta_{upper}^{e} = \alpha_{s} \cdot \hat{\eta}_{upper} + (1 - \alpha_{s}) \cdot \eta_{upper}^{e-1} \\
    \eta_{lower}^{e} = \alpha_{s} \cdot \hat{\eta}_{lower} + (1 - \alpha_{s}) \cdot \eta_{lower}^{e-1}
    \end{aligned}
\right\},
\label{eq:smooth_bound}
\end{equation}
where $\hat{\eta}_{upper}$ and $\hat{\eta}_{lower}$ are computed from \eref{eq:percentile_bound}, and $\alpha_{s} \in (0,1)$ is a smoothing factor controlling the update rate. This mechanism ensures that normalization bounds evolve gradually, mitigating reward instability during training and enabling consistent policy learning as the RUL distribution shifts.

\subsection{Lifespan-guided Reward with ARN}

To incorporate ARN into policy learning, we redefine the \emph{RUL Reward} using the dynamically updated boundaries $\boldsymbol{\eta}_{b}^{e}$ computed for each episode $e$:
\begin{align}
    R_{life}^{(e)}(\eta,\boldsymbol{\eta}_{b}^{e}):= \mathbb{I}\{\text{task completed}\} \cdot
        \frac{\eta-\eta_{lower}^{e}}{\eta_{upper}^{e}-\eta_{lower}^{e}},
\end{align}
where these bounds ensure that reward scaling reflects the agent's evolving performance.

The final \emph{Lifespan-guided Reward} used in training integrates this normalized RUL Reward into the credit-assigned formulation from \eref{eq:credit_assign}:
\begin{align}
\label{eq:DRFP}
    R^{(e)}_{ARN}(s,a,\eta, \boldsymbol{\eta}_{b}^{e}) := R_{task}(s,a) + \gamma_{b}^{T-t} R_{life}^{(e)}(\eta,\boldsymbol{\eta}_{b}^{e}),
\end{align}
where $e$ indexes the episode. This design allows the reward function to adapt over time, aligning learning signals with observed RUL distributions and supporting robust policy optimization.
\begin{figure*}[t]
    \centering
    \includegraphics[width=1\linewidth]{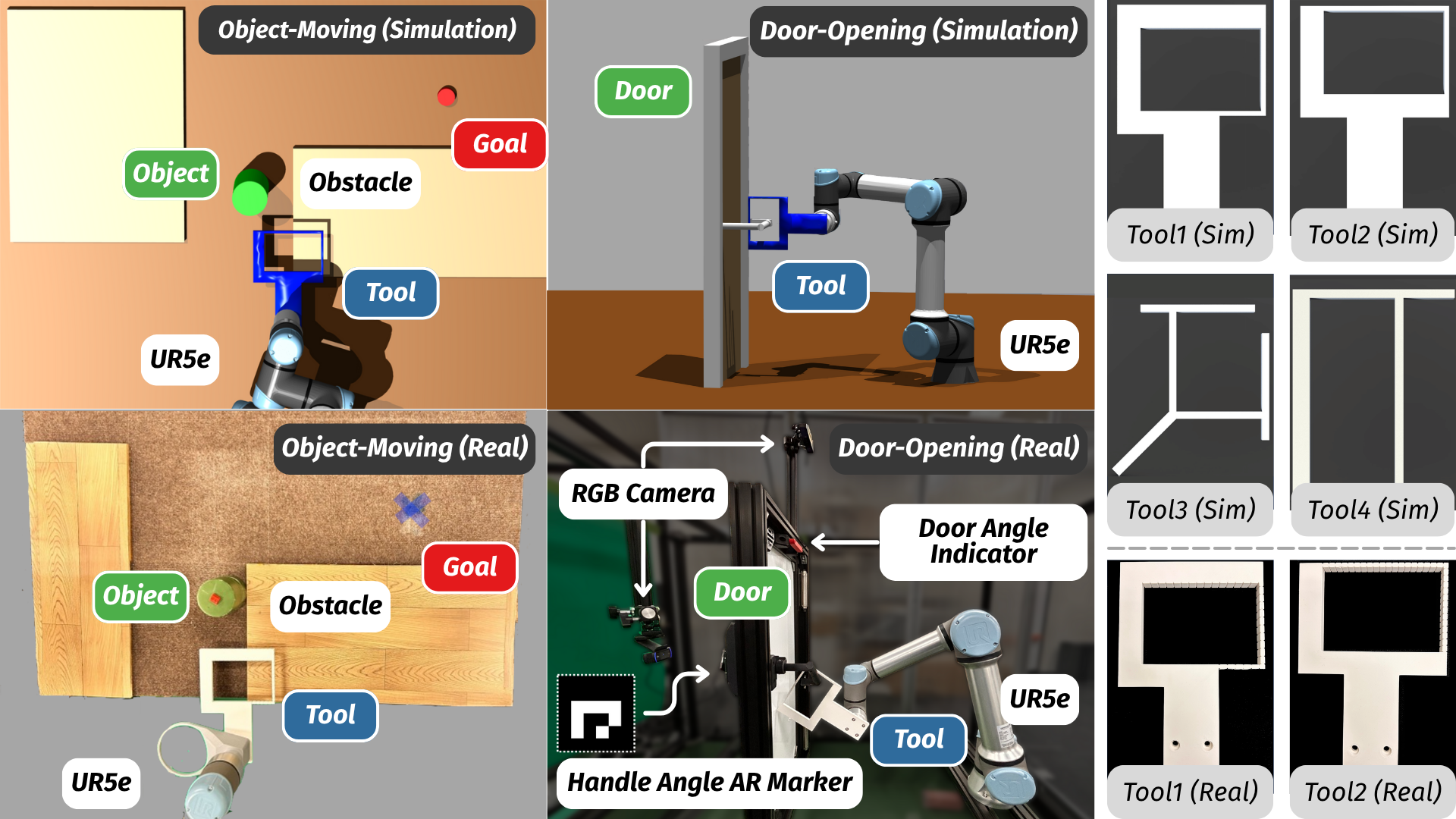}
    \caption{
    \textbf{Overview of the experimental environments and tools used in the study.} The left panels show the Object-Moving and Door-Opening tasks in both simulation and real-world setups. In the Object-Moving task, a UR5e robot with a 3D-printed tool pushes a 2-kg cylindrical object toward a target location on a planar surface, with obstacles and a carpeted floor introducing realistic friction. Object positions are tracked using an Intel RealSense camera via a red dot marker. The middle panels show the Door-Opening task, where the robot uses a 3D-printed tool to press down and rotate a door handle, then pull the door open to 30 degrees. Real-world implementation includes a hinged acrylic door with a mounted handle, and an RGB camera tracks the door angle using an AR marker placed behind the handle. The right panels displays the four tool geometries used in simulation (Tools 1–4) and the two fabricated with ASA filament via 3D printing for real experiments (Tools 1 and 2). To ensure consistent failure patterns and accelerate damage for evaluation, small evenly distributed notches were added to the real tools.
    }
    \label{fig:env_overview}
\end{figure*}

\section{Experiments Setups}\label{sec:Experiments}
\subsection{Key Evaluation Questions}
To assess the effectiveness and generalizability of the proposed method, we conduct a series of simulation and real-world experiments. These evaluations investigate whether the learned policies can adapt general-purpose tool use strategies to prolong general-purpose tool's lifespan while maintaining task success across varied conditions. Our experimental study addresses the following key questions:
\begin{enumerate}
\item [(Q1)] Can the proposed method adapt to different tool shapes and consistently extend their lifespan (evaluated as RUL) compared to a baseline that ignores lifespan?
\item [(Q2)] Does the ARN mechanism improve policy learning stability and lifespan extension compared to a static reward function?
\item [(Q3)] Does incorporating internal stresses, beyond external force measurements, lead to more effective tool use policies for lifespan preservation?
\item [(Q4)] Can the method be effectively applied to the task with richer contacts?
\item [(Q5)] Does the learned policy retain its ability to extend tool lifespan when transferred from simulation to physical robotic systems?
\end{enumerate}
Questions Q1 through Q4 are evaluated through simulation experiments, while Q5 is validated through real-world hardware experiments.

\subsection{Tool use Tasks and Environments}
We designed two representative tasks, Object-Moving and Door-Opening, to evaluate policy learning across varying physical interactions. Both tasks are implemented in the real world and mirrored in simulation using Isaac Gym to enable training and sim-to-real transfer. An overview is shown in \fref{fig:env_overview}.

\subsubsection{Object-Moving Task}
A UR5e robotic arm uses a mounted tool to push a 2-kg cylindrical object to a predefined goal on a planar surface. The object is initialized randomly within a specified region, and a carpeted surface increases friction, introducing realistic resistance. The task is considered successful when the object reaches the target zone. The simulation environment matches the real setup in object geometry, tool configuration, motion constraints, and task conditions. This task serves as a structured but less contact-intensive benchmark for evaluating tool use efficiency and wear reduction.

\subsubsection{Door-Opening Task}
This task requires a UR5e robotic arm to use a tool to press down and rotate a door handle, then pull the door open against a restoring torque applied to both the handle and the door. The robot and tool are initialized in a ready-to-open pose at the start of each episode to isolate tool use strategy learning from handle localization or grasp planning. The real-world environment is constructed with aluminum framing, a hinged acrylic door panel, and a mounted handle. The simulation clone replicates the geometry, dynamics, and constraints to ensure consistent evaluation. Task success is defined by reaching a target door opening angle ($30^\circ$), presenting a more contact-rich scenario to test tool use strategies under complex interactions.

\subsection{Tool Variations}
We designed four general-purpose tool geometries to evaluate policy learning across diverse shapes, as shown in the right column of \fref{fig:env_overview}:
\begin{enumerate}[label=(Tool\arabic*), align=left, leftmargin=*]
\item Broad handle with a rectangular top section, narrower at the bottom and right sides.
\item Broad handle with a rectangular top, narrower at the upper and right sides.
\item Double-T shape with the handle offset from the central axis.
\item Inverted F-shape with a thicker handle region.
\end{enumerate}

In simulation, the Object-Moving task employs all four tools to test generalization across shapes, while the Door-Opening task uses Tool1 and Tool2 for handle interaction. To approximate varying material strength in a single-material FEA model, we adjusted structural thickness across tool regions to create different strength distributions. Given that FEA stress distribution patterns remain consistent when geometry and boundary conditions are fixed, all tools were modeled with ASA material properties, adopting parameters (elastic modulus, shear modulus, Poisson’s ratio, density, and S–N curve) from \cite{brvcic2021asa}.

In real-world experiments, both tasks use Tool1 and Tool2, manufactured using 3D-printed ASA (ApolloX) filament. To focus evaluation on policy effects rather than precise RUL prediction, evenly distributed notches were added to tool surfaces to accelerate failure in a controlled manner. This ensured observed lifespan differences resulted from learned usage strategies rather than random stress concentrations.

\subsection{Policy Learning with Domain Randomization}
Policies for both tasks are trained in simulation and then deployed directly in real-world experiments. Our method adopts the SAC algorithm with domain randomization to improve sim-to-real transfer. During training, parameters such as tool surface smoothness, environmental friction coefficient, and object mass (for Object-Moving) are randomized to expose the policy to diverse conditions reflecting real-world variability. The network architecture follows \cite{kadokawa2023cyclic}.

Reward design follows the task structure described above, integrating both task completion objectives and lifespan-guided considerations from Section~\ref{sec:Proposed_method}. Policies are trained entirely in simulation before deployment on the UR5e without additional fine-tuning. Hyperparameters are listed in Table~\ref{tab:params}.

\subsubsection{Learning in the Object-Moving Task}
In simulation, the UR5e’s end-effector is constrained to planar XY motion at a fixed height. The action space has 2 dimensions representing movements along $x$ and $y$ axes ($\mathcal{A}\in\mathbb{R}^2$). The observation space includes relative positions between the end-effector and object, and between the object and the goal ($\mathcal{S}\in\mathbb{R}^4$).

The task reward $R_{task}^{{obj}}$ encourages efficient pushing toward the target:
\begin{align}
\label{eq:PD_TASK_reward}
R_{task}^{{obj}} = -(d_{x} + d_{y} + \alpha_{dis} \cdot d_{ee}),
\end{align}
where $d_{x}$ and $d_{y}$ are distances to the goal, and $d_{ee}$ is the distance between the end-effector and the object. $\alpha_{dis}$ weights the approach and pushing components.

\subsubsection{Learning in the Door-Opening Task}
Here, the end-effector executes rotation and translation to manipulate the handle. The action space includes rotation around the $x$-axis and translation along $y$ and $z$ axes ($\mathcal{A}\in\mathbb{R}^3$). The observation space includes the end-effector’s rotation angle and position, handle rotation angle, and door opening angle ($\mathcal{S}\in\mathbb{R}^5$).

The reward function incentivizes maximizing the door’s opening angle relative to the target:
\begin{align}
\label{eq:OD_TASK_reward}
R_{task}^{{door}} = \frac{\theta_{door}}{\theta_{target}},
\end{align}
where $\theta_{door}$ is the current door angle and $\theta_{target}$ is the desired target angle. This formulation aligns with the success criteria defined for the Door-Opening task, guiding the agent to learn effective pressing and pulling strategies.

\subsection{Baselines for Comparison}
This section describes the baselines for validating the effectiveness of our method. Along with our proposed approach, we designed three baselines with different reward functions that consider different metrics:
\begin{itemize}
    \item \emph{\textbf{Ours}}: The proposed lifespan-guided tool use reinforcement learning.
    \item \emph{\textbf{Baseline}}: This approach only considers the task goal of moving the object to the target point. Therefore, the reward function is defined by \eref{eq:PD_TASK_reward} and ~\eref{eq:OD_TASK_reward}
    \item \emph{\textbf{Ours w/o ARN}}: As an ablation study for the proposed ARN method, a static reward function is employed to formulate lifespan rewards, with the objective of maximizing them: $R_{\hat\eta} = -\frac{1}{\eta}+1$.
    This reward incentivizes longer lifespans, which increases monotonically and approaches 1 as lifespan grows.
    \item  \emph{\textbf{Torque}}: A torque reward function is designed to minimize the torque $\tau$ at the end-effector. This method also using the ARN approach, but replace RUL with torque. Since torque should be minimized, the reward is defined as one minus the normalized $\tau$ value: $R_{\tau} = 1 - \frac{(\tau - \tau_{\min})}{\tau_{max}-\tau_{min}}.$
\end{itemize}
\begin{figure}[t]
    \centering
    \includegraphics[width=1\hsize]{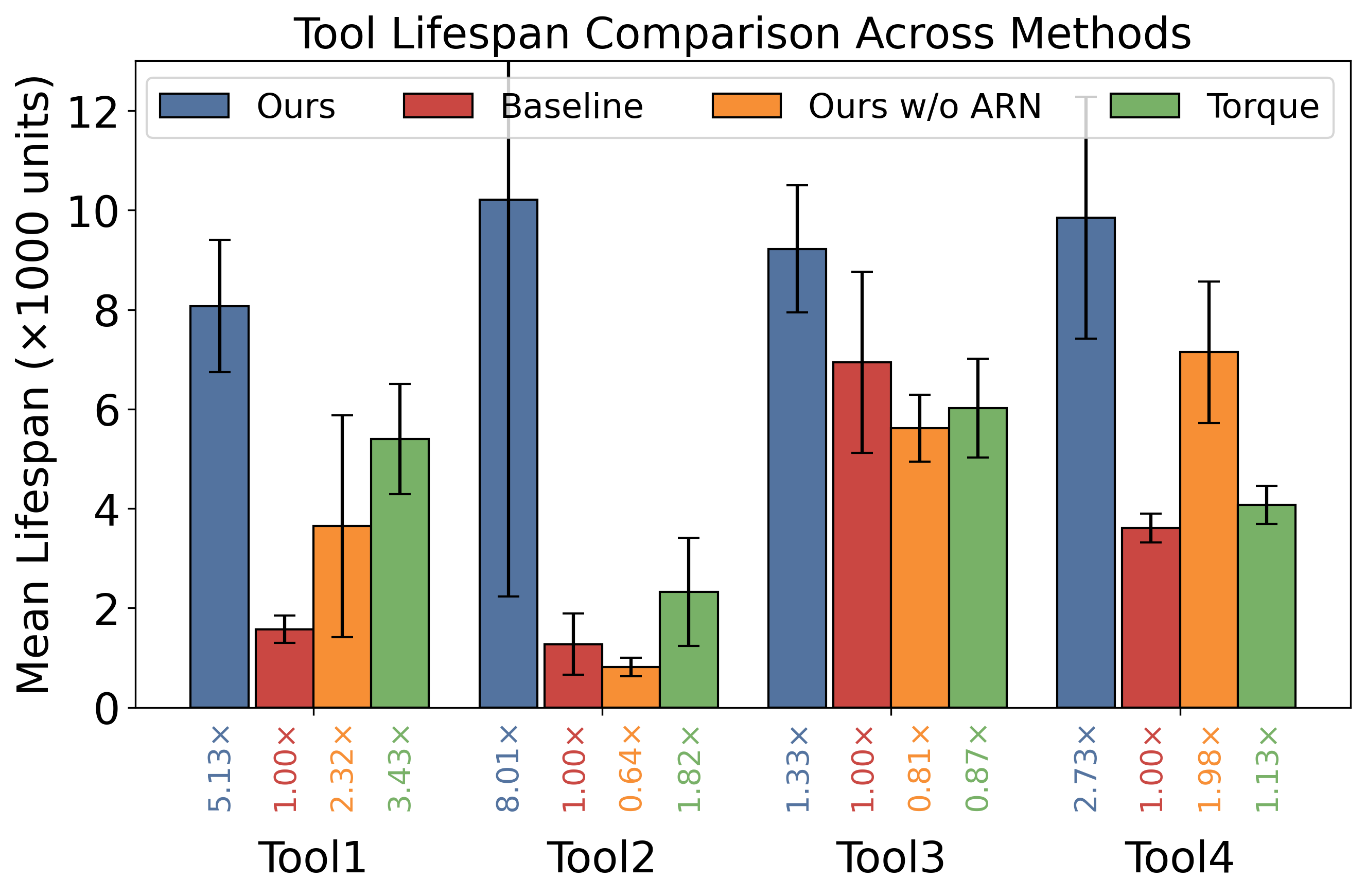}
    \caption{
    \textbf{Simulation results of the Object-Moving experiment comparing tool lifespan across methods.} Bars represent the mean RUL averaged over 100 trials for four policies: \emph{\textbf{Ours}}, \emph{\textbf{Baseline}}, \emph{\textbf{Ours w/o ARN}}, and a Torque-minimizing policy. Error bars indicate standard deviation. The multiplier values below each bar (e.g., $5.13\times$) denote the improvement relative to the Baseline. Results demonstrate that \emph{\textbf{Ours}} consistently achieves longer tool lifespan across all tool shapes, confirming its effectiveness and generalization capability.
    }
    \label{fig:sim_result}
\end{figure}

\begin{figure}[t]
    \centering
    \includegraphics[width=1\hsize]{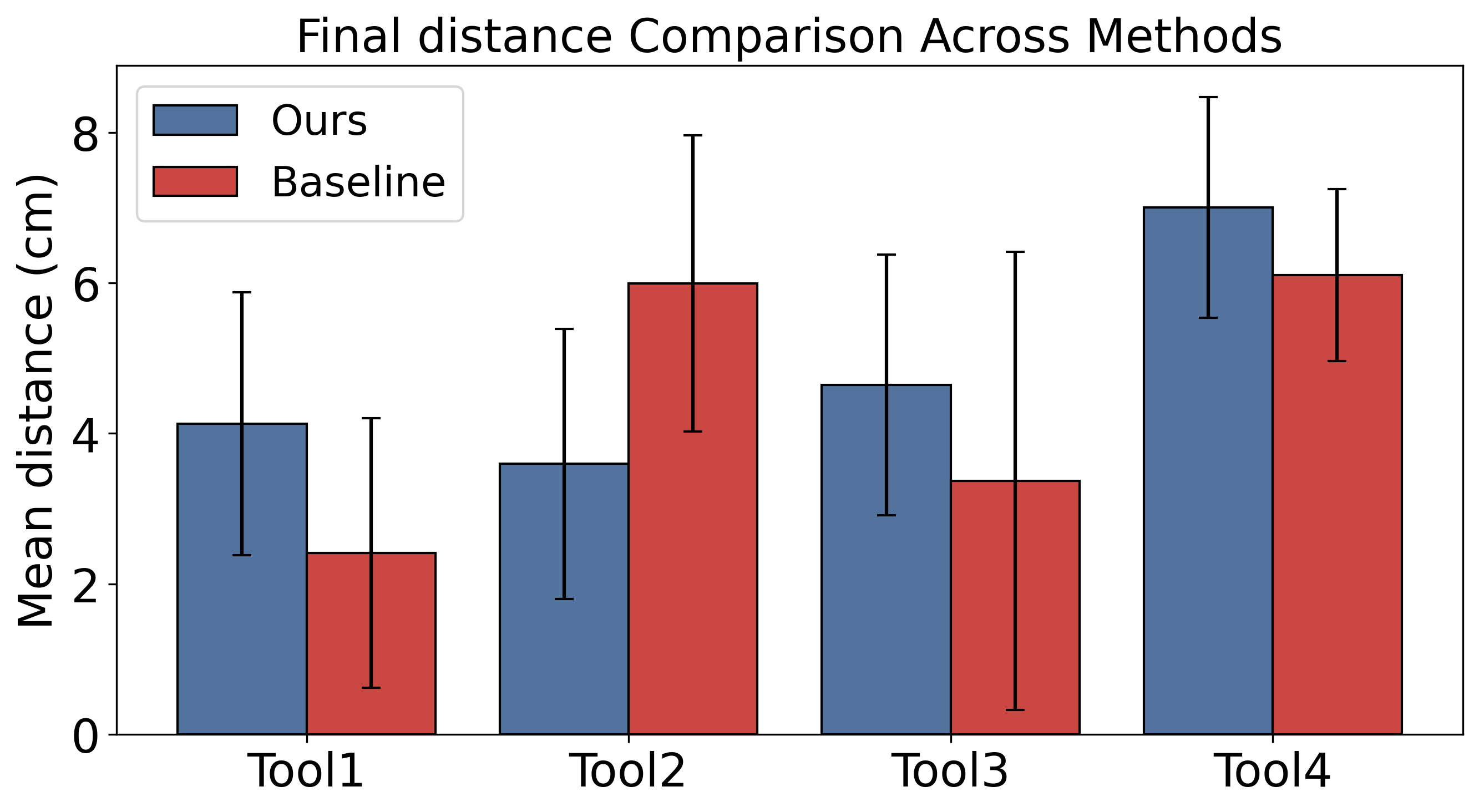}
    \caption{
    \textbf{Comparison of task performance in the Object-Moving experiment.} The bar chart shows the final mean distance between the object and the goal across trials, with shorter distances indicating better task completion performance. Error bars represent standard deviation. Results demonstrate that our method achieves a comparable level of task completion accuracy to the Baseline, confirming that lifespan-guided optimization does not compromise task success.
    }
    \label{fig:sim_task_performance}
\end{figure}

\begin{figure*}[t]
    \centering
    \begin{minipage}{0.47\linewidth}
        \includegraphics[width=\linewidth]{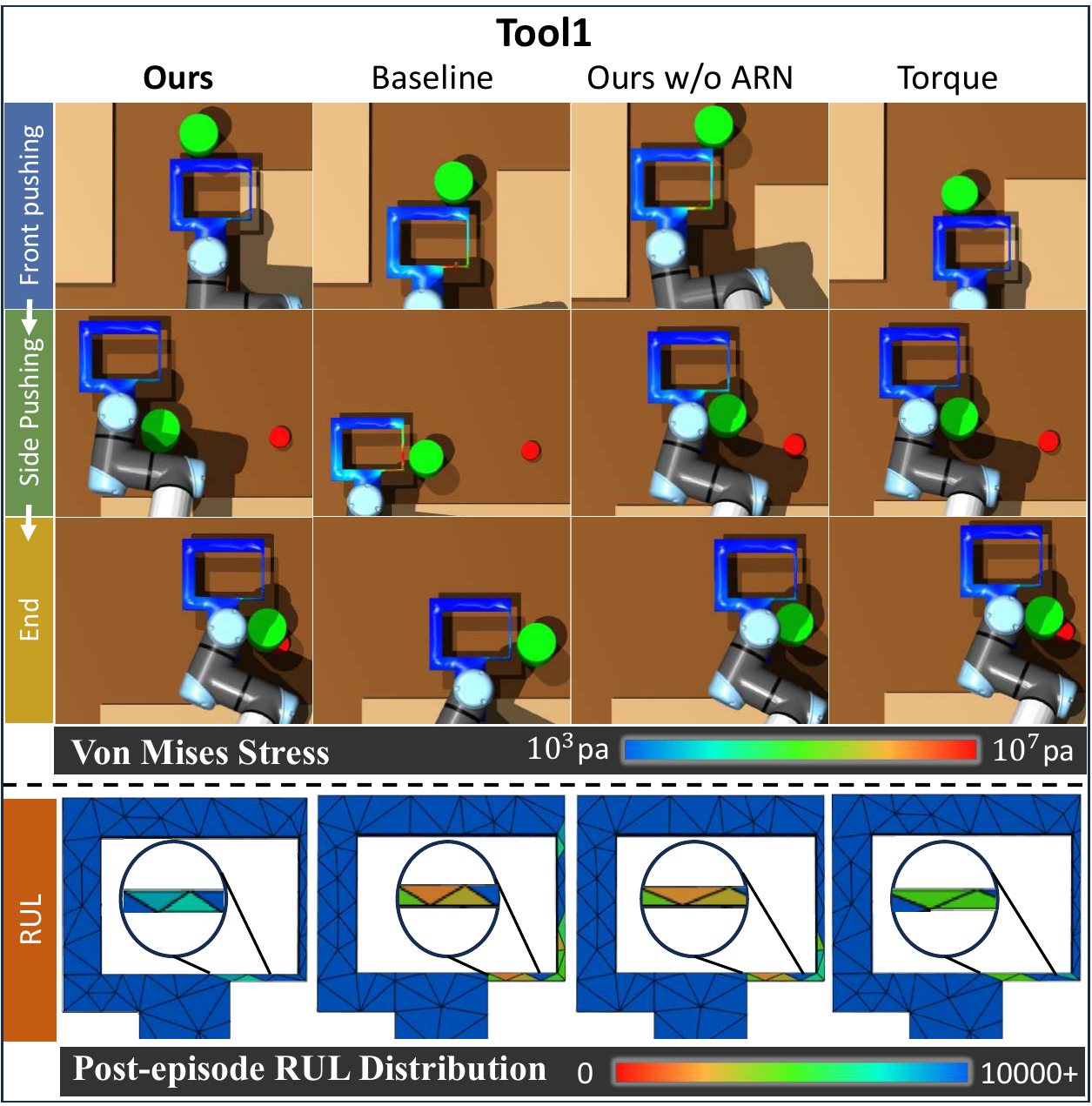}
    \end{minipage}
    \begin{minipage}{0.47\linewidth}
        \includegraphics[width=\linewidth]{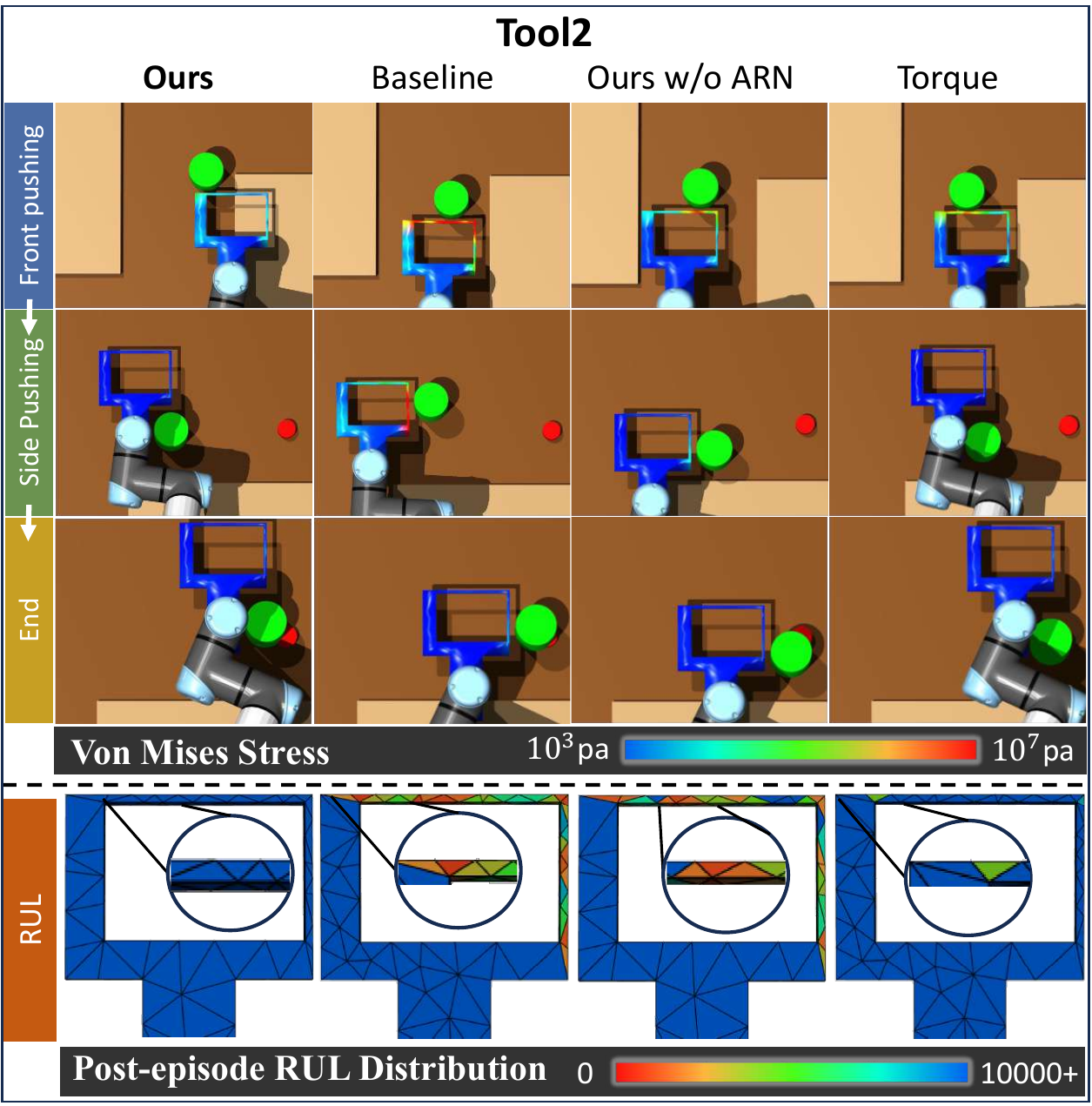}
    \end{minipage}
    \begin{minipage}{0.47\linewidth}
        \includegraphics[width=\linewidth]{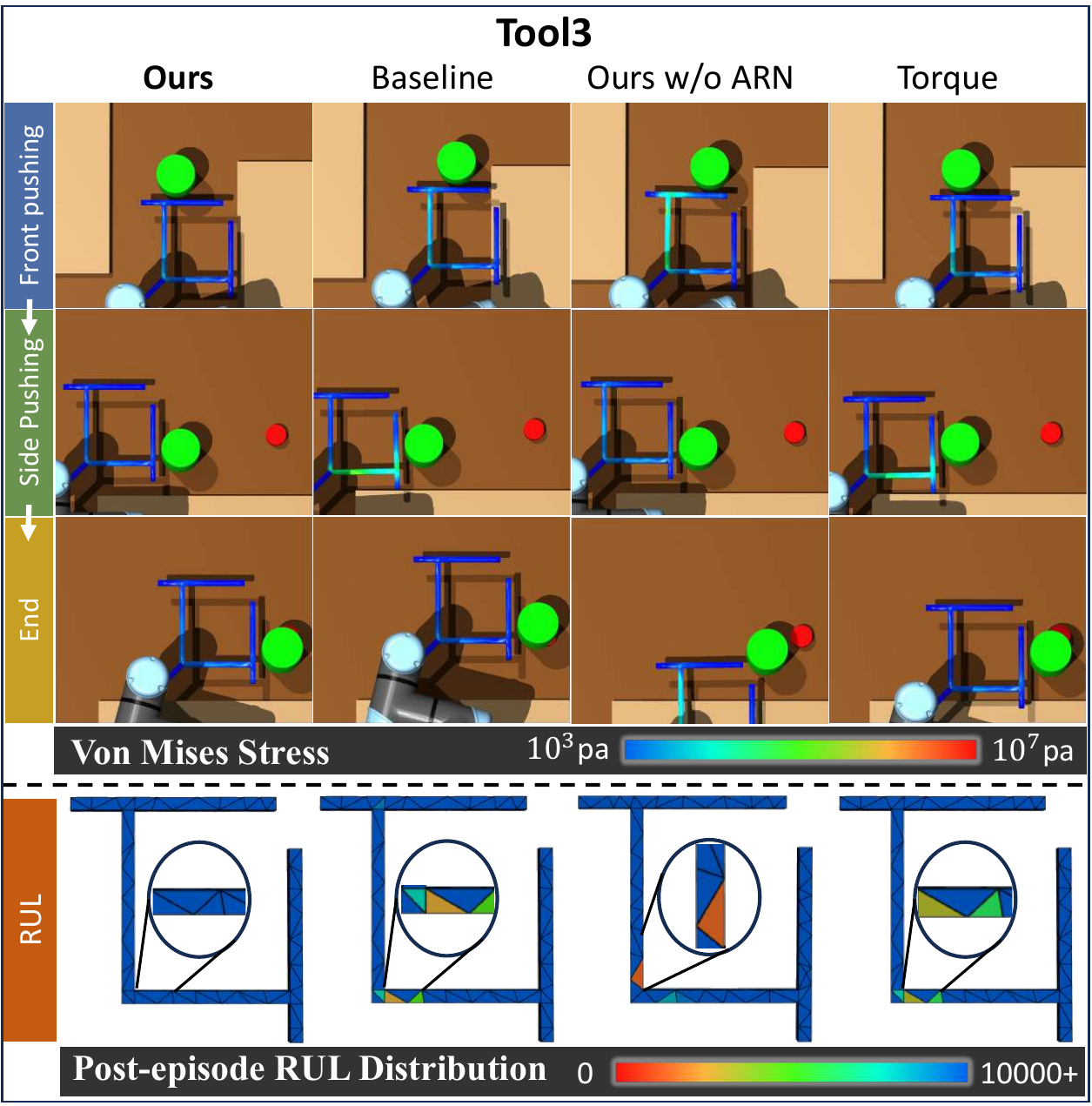}
    \end{minipage}
    \begin{minipage}{0.47\linewidth}
        \includegraphics[width=\linewidth]{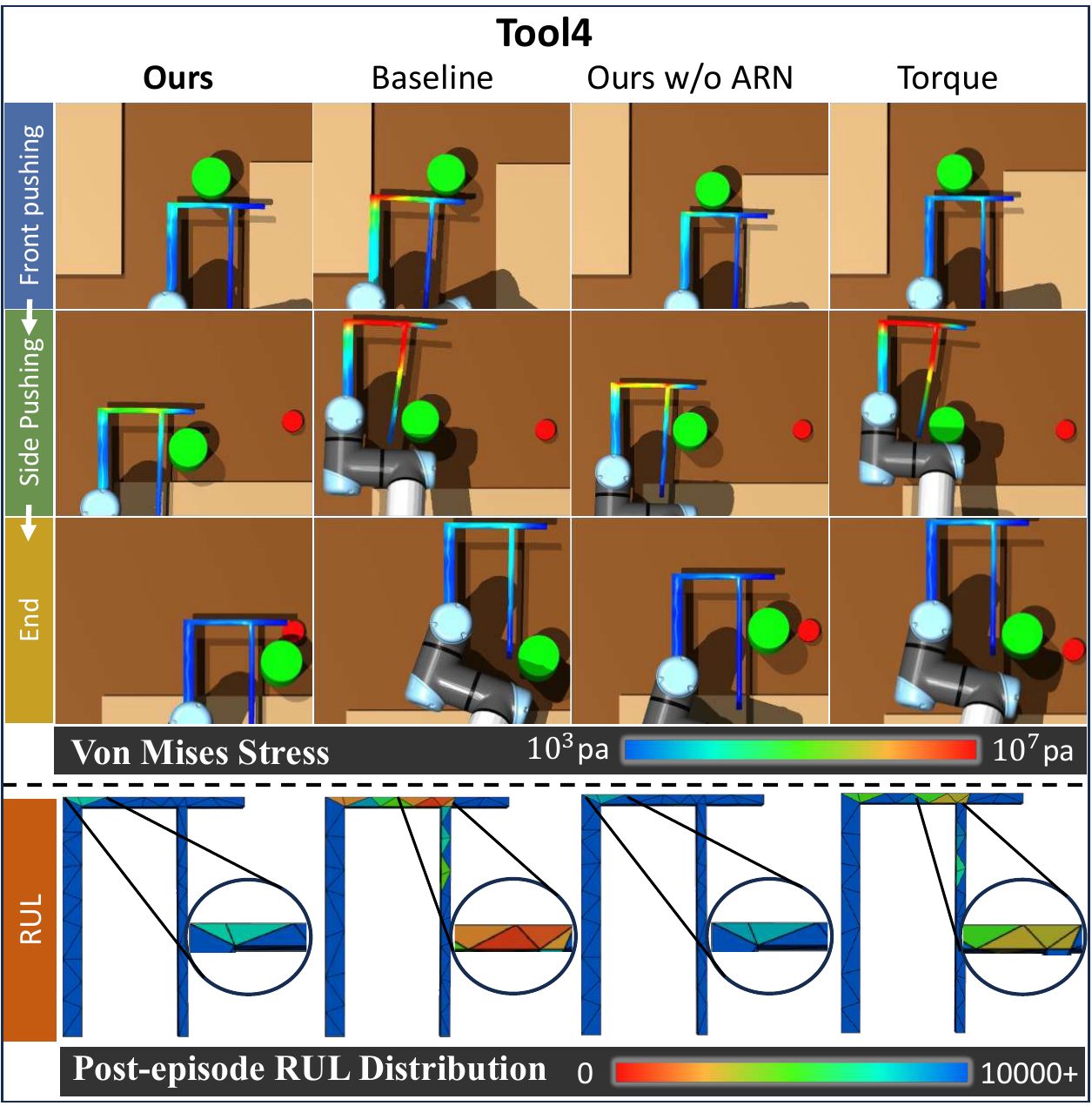}
    \end{minipage}
    \caption{ 
    \textbf{Visual analysis of tool use strategies, stress distribution, and RUL in the Object-Moving task.} For each policy (column), the top two rows show the key front pushing and side pushing phases, with color overlays indicating von Mises stress on the tool (blue indicating low stress, yellow to red indicating high stress). The bottom row presents the resulting RUL heatmap, with values mapped to the color scale below. \emph{\textbf{Ours}} consistently learns to use structurally stronger regions by selecting contact points that minimize internal stress during both phases, leading to higher RUL. In contrast, baseline policies often choose contact regions that are easier for task completion but structurally weaker, concentrating stress and reducing lifespan.
    }
    \label{fig:Trajetories}
\end{figure*}

\section{EXPERIMENT RESULTS}
In this section, we address five research questions (Q1–Q5) through both simulation and real-robot experiments. Demonstration videos are available on our project page: https://sites.google.com/view/lifespan-guided-rl/.
\qsubsection{1}{Can the proposed method adapt to different tool shapes and consistently extend their lifespan (evaluated as RUL) compared to a baseline that ignores lifespan?}
\subsubsection{Overview}
To address Q1, we evaluated whether the proposed method can adapt to different tool shapes and consistently extend their lifespan when compared to a baseline that ignores lifespan considerations. Experiments were conducted in the simulated Object-Moving task using all four tool designs introduced in \fref{fig:env_overview}. Specifically, \emph{\textbf{Ours}} framework was compared against a \emph{\textbf{Baseline}} that optimizes only for task completion, with its reward function defined in \eref{eq:PD_TASK_reward}. This setup allows us to assess \emph{\textbf{Ours}}’s ability to learn task-completing policies that explicitly account for structural damage across varied tool geometries.

\subsubsection{Results}
We evaluated policies learned with both the \emph{\textbf{Baseline}} and \emph{\textbf{Ours}} approaches across 100 trials per tool, randomizing the initial object position and environment parameters at each episode. \fref{fig:sim_result} summarizes the mean RUL for all four tool designs. \emph{\textbf{Ours}} consistently achieved higher RUL than the \emph{\textbf{Baseline}}, with improvements ranging from $1.33\times$ to $8.01\times$. These results demonstrate that explicitly incorporating lifespan estimation into policy learning enables the agent to significantly extend tool lifespan, validating \emph{\textbf{Ours}} effectiveness in learning tool use strategies that consider structural damage.

\fref{fig:sim_task_performance} compares task execution performance by reporting the final mean Euclidean distance between the object and the target for both methods. The results indicate that \emph{\textbf{Ours}} achieves task performance comparable to the \emph{\textbf{Baseline}}, with only minor performance trade-off despite prioritizing lifespan extension. This finding confirms that the proposed approach can balance task-completing policy objectives with lifespan-guided considerations without sacrificing overall task success.

To further analyze how the lifespan reward shapes tool use strategies, \fref{fig:Trajetories} illustrates key front pushing and side pushing phases along with \emph{von Mises stress distributions} and resulting RUL maps. Across all four tools, \emph{\textbf{Ours}} learns to utilize structurally stronger regions, minimizing internal stress and avoiding vulnerable points. For example, while the \emph{\textbf{Baseline}} often favors upper-right contact areas (Tool1 and Tool2) that are easier for task completion but prone to stress accumulation, \emph{\textbf{Ours}} adopts trajectories that shift contact toward more robust sections. Similar patterns are observed with Tool3 and Tool4, where \emph{\textbf{Ours}} consistently adapts contact strategies to reduce stress fluctuations and extend lifespan, highlighting the benefit of incorporating lifespan-guided reasoning into the learned policy.

\subsubsection{Summary}
These results directly address Q1 by demonstrating that our method generalizes across diverse tool geometries and consistently extends lifespan compared to the baseline. By learning to adapt contact strategies based on each tool’s structural characteristics, the agent achieves significant improvements in RUL across all four designs. This validates that the proposed approach is not limited to a single shape but is effective at discovering lifespan-guided tool use policies for varied general-purpose tool geometries, confirming its ability to handle different tool shapes while maintaining task performance.

\begin{figure}[t]
    \centering
    \includegraphics[width=1\hsize]{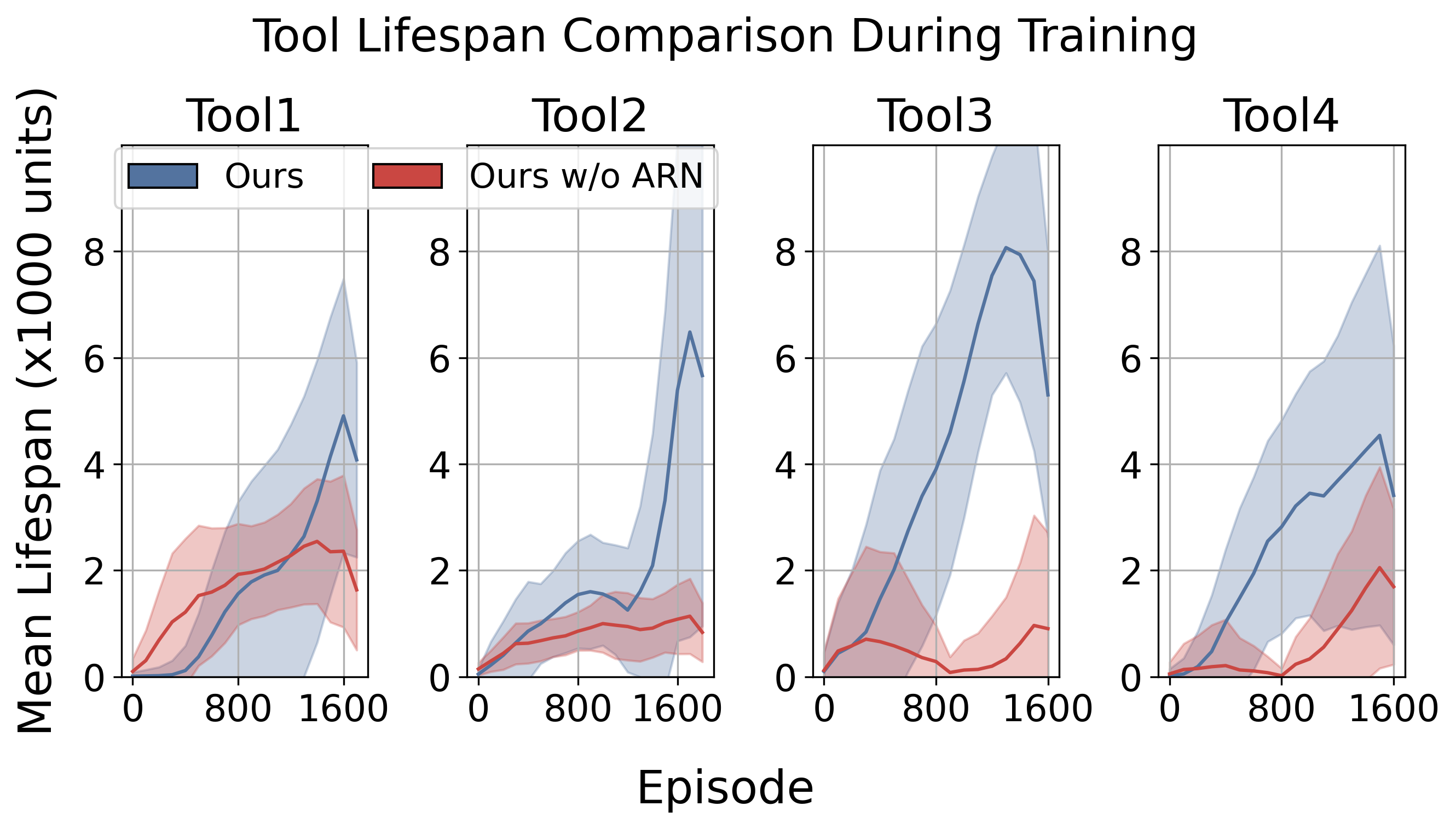}
    \caption{
    \textbf{Comparison of tools' lifespan during training with and without ARN.} Each subplot shows the RUL (mean in solid line with standard deviation in shades) across training episodes of each tool. The results demonstrate that ARN significantly improves training stability that promotes consistent lifespan extension across different tools.
    }
    \label{fig:training_lifespan}
\end{figure}

\qsubsection{2}{Does the ARN mechanism improve policy learning stability and lifespan extension compared to a static reward function?}
\subsubsection{Overview}
To investigate Q2, we conducted ablation studies in the Object-Moving simulation task to evaluate the impact of the Adaptive Reward Normalization (ARN) mechanism. We compared our proposed \emph{\textbf{Ours}} framework with \emph{\textbf{Ours w/o ARN}}, which replaces adaptive normalization with a static lifespan reward function. Both methods were tested on all four tool designs. This comparison allows us to isolate the effect of ARN on policy learning stability in extending a tool's lifespan.

\subsubsection{Results}
Comparing tool lifespan outcomes in \fref{fig:sim_result}, we observe that \emph{\textbf{Ours}} consistently achieves longer mean RUL than \emph{\textbf{Ours w/o ARN}} across all four tools. To investigate the reason for this improvement, \fref{fig:training_lifespan} plots the evolution of mean RUL during training. The results show that \emph{\textbf{Ours}} exhibits a steady and substantial increase in lifespan over episodes, while \emph{\textbf{Ours w/o ARN}} demonstrates limited improvement. This indicates that ARN enables more stable exploration of higher-RUL strategies throughout training.

To further validate this behavior, \fref{fig:Trajetories} provides a qualitative comparison of learned tool-use trajectories and resulting stress distributions. \emph{\textbf{Ours}} consistently selects contact regions that minimize internal stress, while \emph{\textbf{Ours w/o ARN}} demonstrates a limited ability to identify and exploit such lower-stress regions. This difference demonstrates that adaptive reward normalization guides the agent toward safer, lower-damage usage patterns, supporting effective lifespan extension.

\subsubsection{Summary}
These results provide clear evidence addressing Q2 by demonstrating that the ARN mechanism significantly improves both policy learning stability in extending tool lifespan compared to using a static reward function. By adaptively scaling rewards based on observed RUL distributions, ARN addresses the challenge of unknown lifespan ranges before execution, ensuring appropriate reward scaling as learning progresses. This not only ensures consistent learning of tool use strategies that generalize across different shapes, but also enables the agent to more reliably explore and discover policies achieving higher RUL. Overall, the results validate that ARN is essential for effectively incorporating lifespan considerations into policy optimization.

\qsubsection{3}{Does incorporating internal stresses, beyond external force measurements, lead to more effective tool use policies for lifespan preservation?}
\subsubsection{Overview}
To address Q3, we evaluated whether explicitly incorporating internal stresses, rather than relying solely on external force measurements, leads to more effective tool use policies for lifespan preservation. We compared our proposed \emph{\textbf{Ours}} framework with \emph{\textbf{Torque}}, which optimizes policies by minimizing end-effector torque without considering internal stress. Both methods were evaluated using the same Object-Moving simulation setup as in Q1.

\subsubsection{Results}
\fref{fig:sim_result} shows that while the \emph{\textbf{Torque}} method achieves some RUL improvement on Tool1, Tool3, and Tool4, \emph{\textbf{Ours}} consistently delivers superior and more balanced performance across all tools. Notably, \emph{\textbf{Torque}} performs poorly on Tool2 and significantly underperforms on Tool4 compared to \emph{\textbf{Ours}}. This discrepancy highlights a fundamental limitation: minimizing external torque does not necessarily reduce internal structural stress. In practice, low-torque contact regions can still induce high internal stress concentrations, depending on the tool’s geometry.

Trajectory analyses in Fig.~\ref{fig:Trajetories} further illustrate this issue. The \emph{\textbf{Torque}} baseline consistently selects contact points that minimize operational torque but fails to consider internal stress distribution. For example, during front pushing with Tool2 and side pushing with Tool4, the chosen low-torque regions correspond to areas of maximum internal stress, leading to stress concentration and accelerated damage. 

\subsubsection{Summary}
These results clearly address Q3 by demonstrating that explicitly incorporating internal stress measurements leads to more effective and robust tool use policies for lifespan preservation. Unlike methods that rely solely on minimizing external torque, our approach accounts for internal stress distributions that depend on tool geometry, enabling the agent to avoid high-stress contact regions. This results in consistently extended lifespan across different tools, validating the importance of modeling internal stresses to achieve reliable lifespan-guided policy learning.

\begin{figure}[t]
    \centering
    \includegraphics[width=1\hsize]{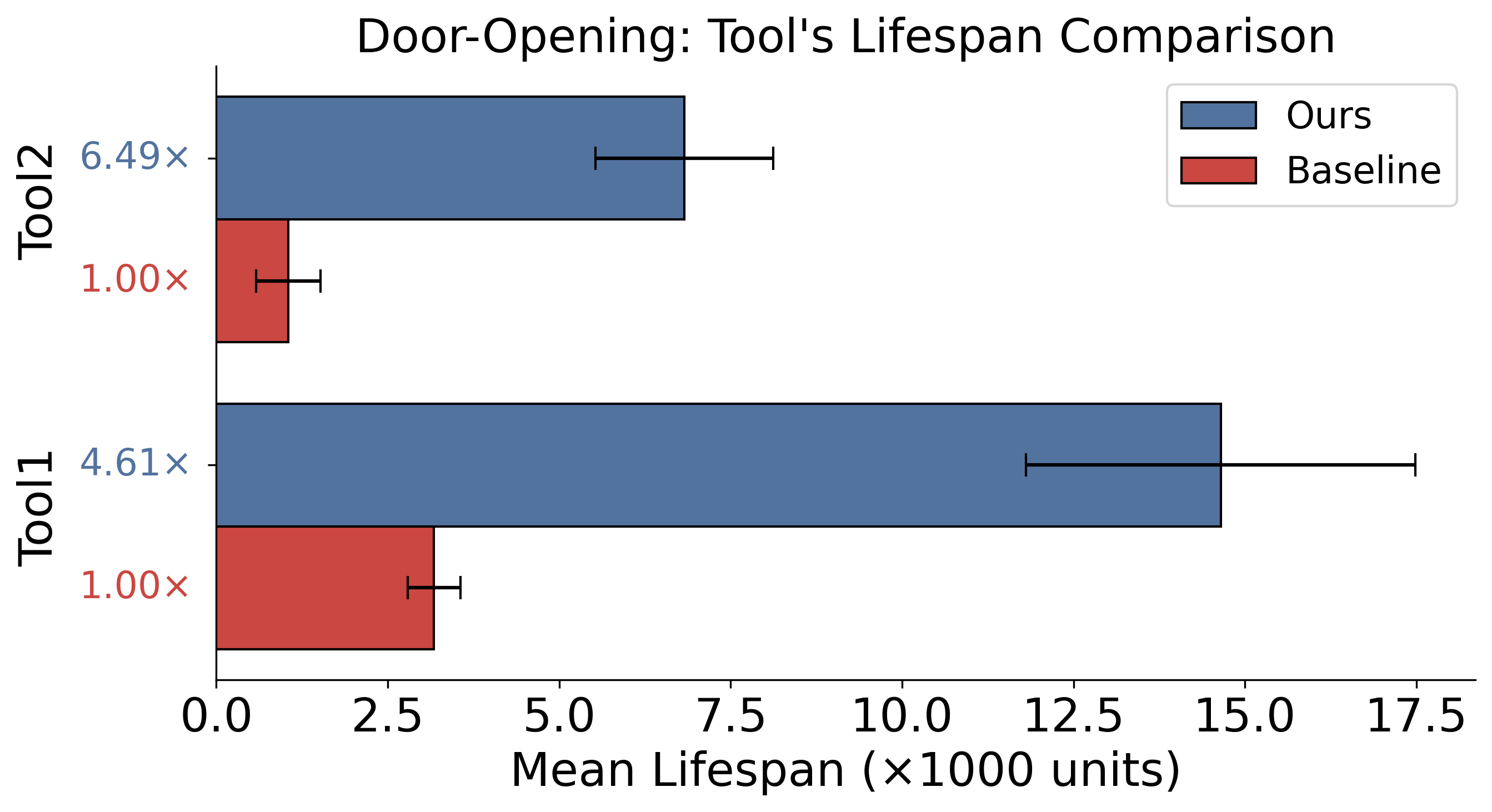}
    \caption{
    \textbf{Simulation results of the Door-Opening experiment comparing tool lifespan across methods.} Bars represent the mean RUL and standard deviation averaged over 100 independent trials for \emph{\textbf{Ours}} and the Baseline. The improvement factors shown beside the y-axis indicate the lifespan gains achieved by \emph{\textbf{Ours}} relative to the Baseline. Results demonstrate that \emph{\textbf{Ours}} substantially increases tool lifespan in the Door-Opening task, confirming its effectiveness in learning strategies that reduce structural damage.
    }
    \label{fig:Sim_result_door}
\end{figure}

\begin{figure}[t]
    \centering
    \includegraphics[width=1\hsize]{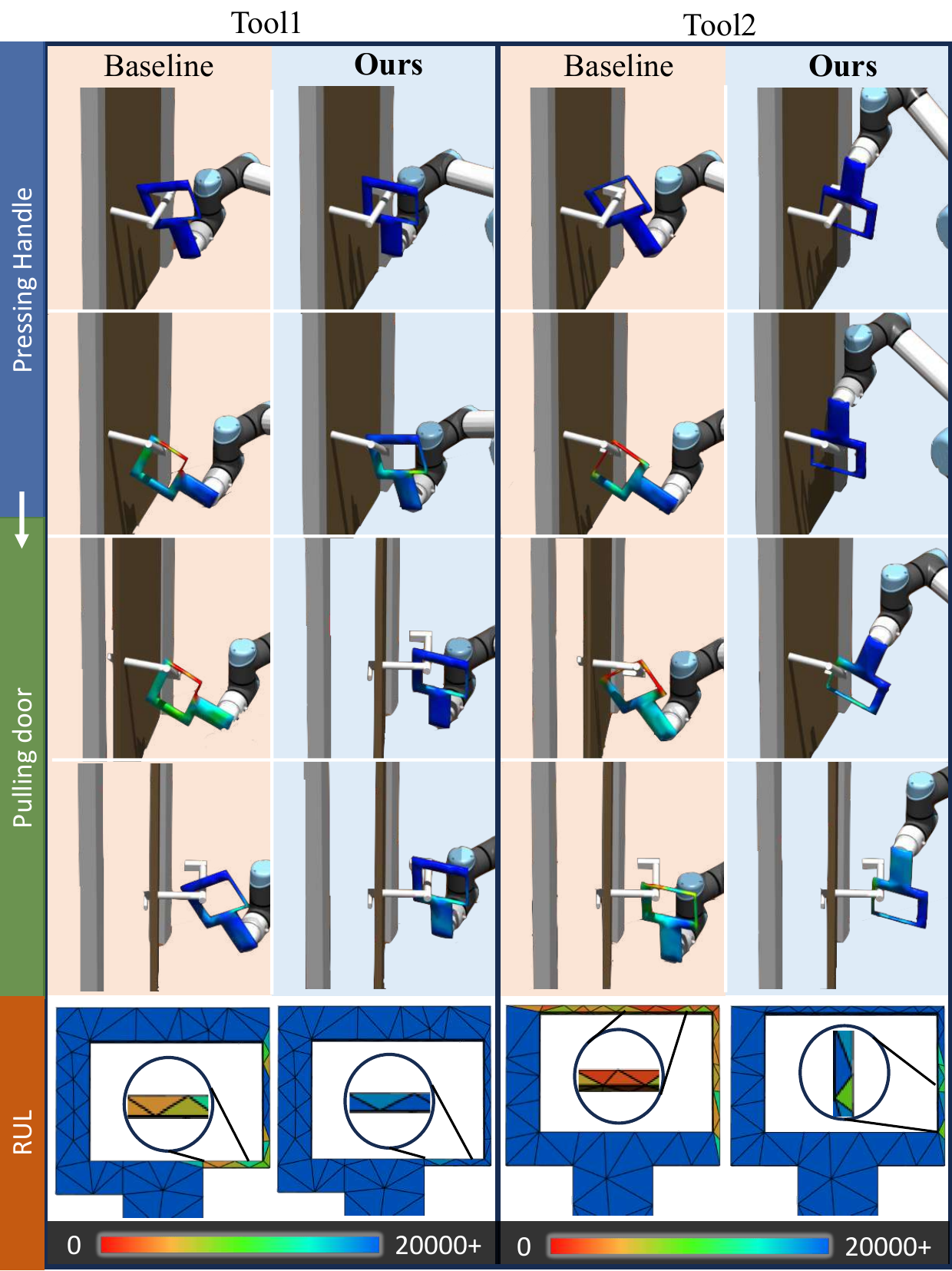}
    \caption{
    \textbf{Visual analysis of tool use strategies, stress distribution, and RUL in the Door-Opening task.} The figure is divided into two phases: ‘Pressing Handle’ and ‘Pulling Door’. The top rows show representative tool use actions with color overlays indicating von Mises stress distribution (blue for low stress, red for high stress). The bottom row displays the resulting RUL heatmaps, with values corresponding to the provided color scale. \emph{\textbf{Ours}} consistently learns to align structurally stronger tool regions with the handle and door motion, minimizing internal stress and extending lifespan. In contrast, baseline policies often apply force to structurally weaker regions, leading to stress concentration and reduced RUL.
    }
    \label{fig:Door_traj}
\end{figure}

\qsubsection{4}{Can the method be effectively applied to the task with richer contacts?}
\subsubsection{Overview}
To address Q4, we evaluate whether the proposed \emph{\textbf{Ours}} framework can be effectively applied to a more complex task with higher contact variability. We focus on the Door-Opening simulation, which features multi-phase interactions, specifically pressing the handle followed by pulling the door, requiring careful control of contact forces and tool orientation. Both \emph{\textbf{Ours}} and the \emph{\textbf{Baseline}} (which optimizes only for task completion) were evaluated under the same experimental setup, using Tool1 and Tool2 as defined in our earlier sections.

\subsubsection{Results}
As summarized in \fref{fig:Sim_result_door}, \emph{\textbf{Ours}} significantly prolongs the tool lifespan compared to the \emph{\textbf{Baseline}}, achieving improvement factors of $4.61\times$ on Tool1 and $6.49\times$ on Tool2. This demonstrates \emph{\textbf{Ours}} ability to generalize lifespan-guided strategies to tasks with complex, contact-rich dynamics.

\fref{fig:Door_traj} illustrates representative trajectories divided into pressing and pulling phases, along with stress distribution and resulting RUL maps. For Tool1, \emph{\textbf{Ours}} aligns the stronger region of the tool with the handle and applies vertical pressing, reducing stress and structural damage. The \emph{\textbf{Baseline}} instead rotates and hooks the handle, inducing wear on weaker areas. During the pulling phase, the baseline maintains this suboptimal posture, exacerbating damage. For Tool2, although the internal shape is similar to Tool1, \emph{\textbf{Ours}} adapts by aligning the bottom section for pressing and pulling, choosing structurally robust regions. 

Although the comparison of task performance was not included, both ours and baseline consistently achieved opening the door to the target angle ($30^\circ$), resulting $100\%$ success rate on both settings.
\subsubsection{Summary}
These results directly address Q4 by confirming that the proposed method effectively generalizes to more complex, contact-rich tasks. By learning to adapt contact strategies across multi-phase interactions, the agent consistently reduces internal stress and extends tool lifespan even under higher variability. This demonstrates that our approach is not limited to simple manipulation tasks but can handle the challenges of complex scenarios requiring precise force control and tool alignment.

\qsubsection{5}{Does the learned policy retain its ability to extend tool lifespan when transferred from simulation to physical robotic systems?}
\subsubsection{Overview}
To address Q5, we evaluated whether the policies learned in simulation retain their ability to extend tool lifespan when directly transferred to physical robotic systems. We conducted real-world experiments that closely replicate the simulation setups for the Object-Moving and Door-Opening tasks. Using 3D-printed Tool1 and Tool2, we compared the performance of our proposed \emph{\textbf{Ours}} framework against the \emph{\textbf{Baseline}} method. For each tool-method combination, we performed three independent trials and recorded the number of successful executions until structural failure to evaluate each tool's actual RUL.

\subsubsection{Results}
\begin{figure}[t]
    \centering
    \begin{minipage}{0.49\linewidth}
        \centering
        \includegraphics[width=\linewidth]{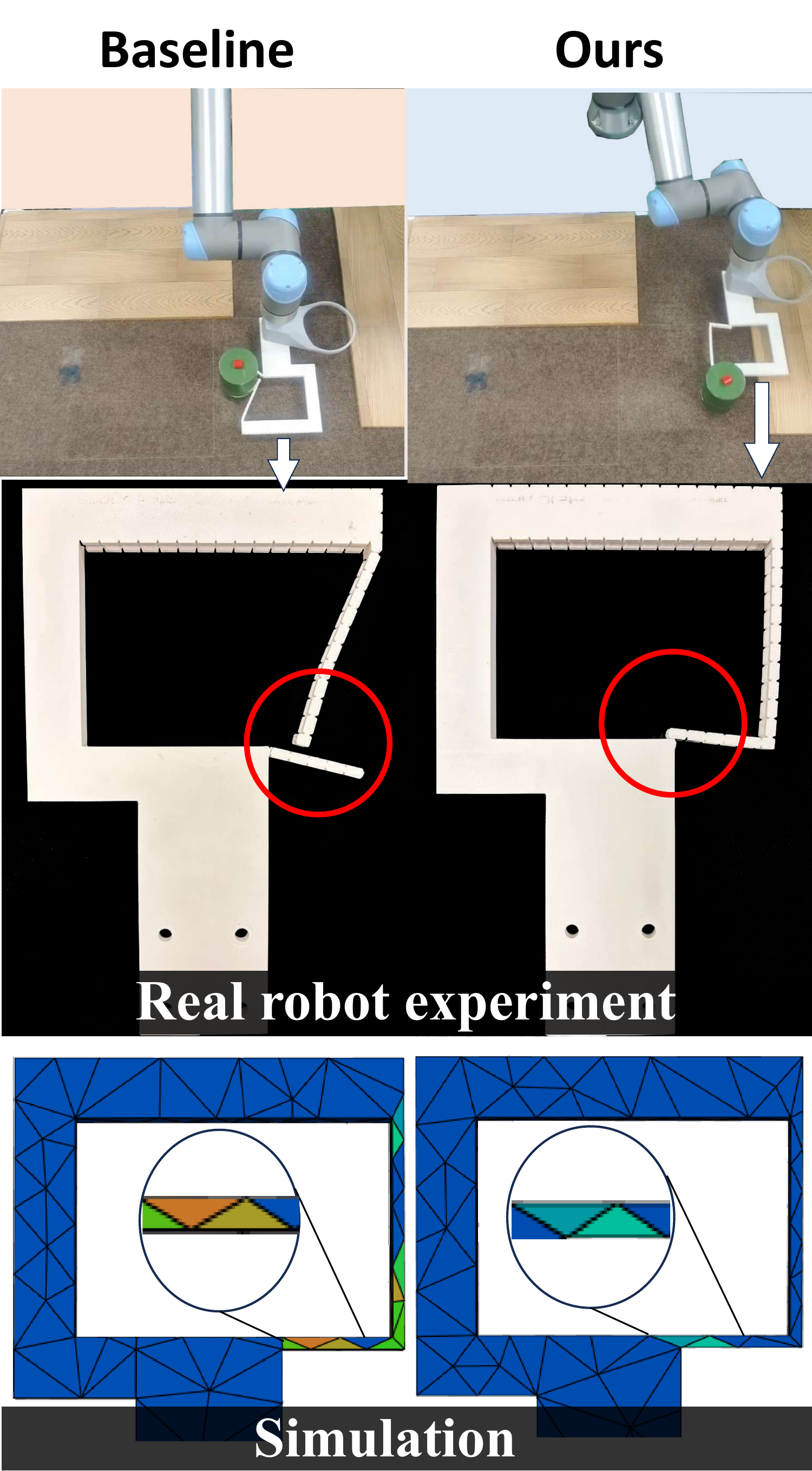}
        \small (a) Tool1
        \label{fig:PD_BT_T1}
    \end{minipage}
    \hfill
    \begin{minipage}{0.49\linewidth}
        \centering
        \includegraphics[width=\linewidth]{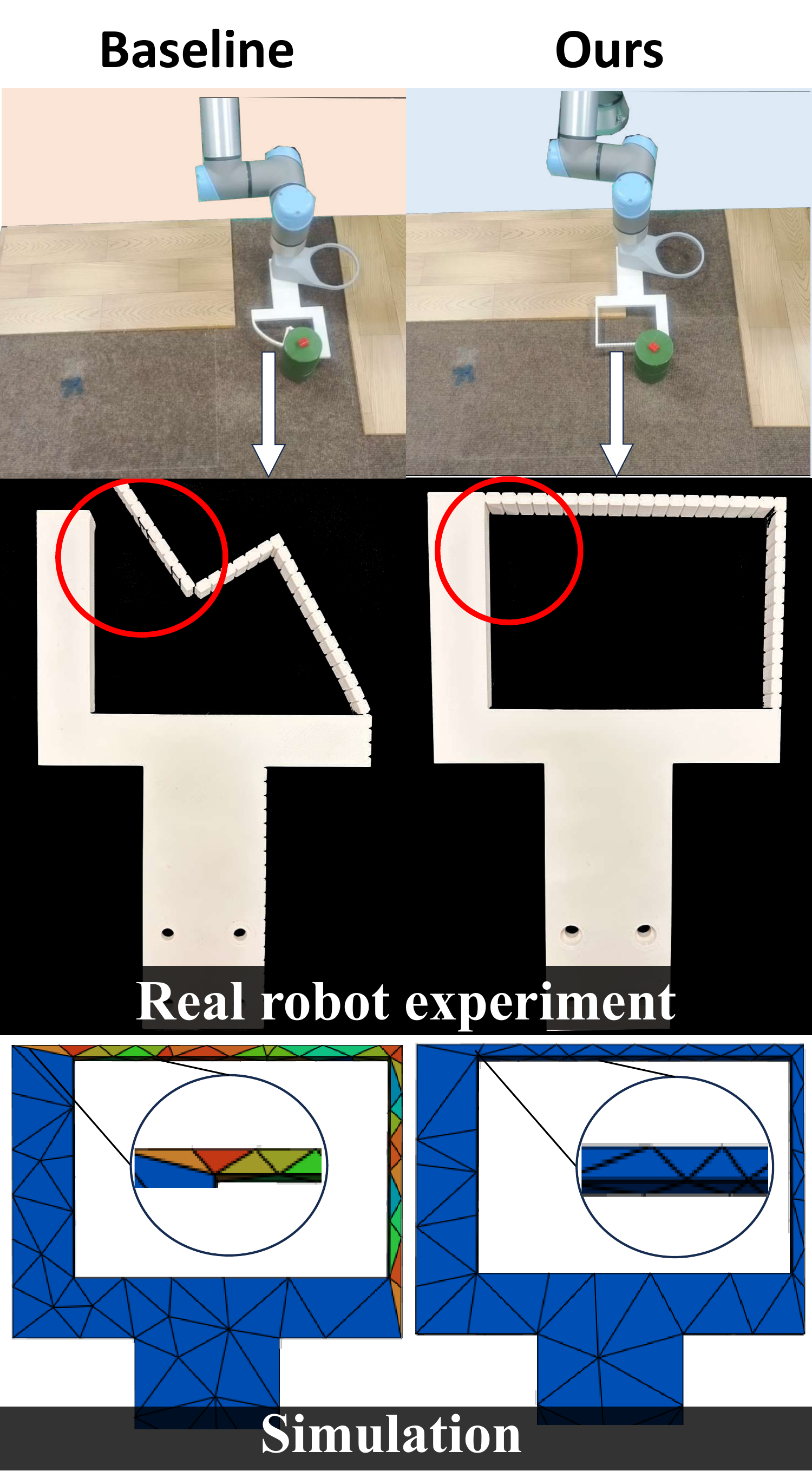}
        \small (b) Tool2
        \label{fig:PD_BT_T2}
    \end{minipage}
    \caption{
    \textbf{Broken tools observed in the Object-Moving experiment, with corresponding simulation analysis.} The top row shows real-world experimental results highlighting fracture locations after repeated use, with red circles indicating failure points. The bottom row presents RUL heatmaps from simulation, identifying regions with minimum predicted lifespan. Comparison confirms that fracture locations in physical experiments align closely with the regions of lowest RUL identified in simulation analysis.
    }
    \label{fig:Broken_Tool_PD}
\end{figure}

Table~\ref{table:Real_result} presents the mean number of executions before failure, along with standard deviations. In the Object-Moving task, \emph{\textbf{Ours}} substantially improved tool durability, achieving over $3\times$ the number of executions for Tool2 and approximately $2\times$ for Tool1 compared to the \emph{\textbf{Baseline}}. \fref{fig:Broken_Tool_PD} shows the resulting fractures: while both methods exhibited failure at the lower rod connection in Tool1, Tool2 fractured at the upper rod center in the \emph{\textbf{Baseline}} case but showed no significant damage under \emph{\textbf{Ours}}. 

\begin{figure}[t]
    \centering
    \begin{minipage}{0.49\linewidth}
        \centering
        \includegraphics[width=\linewidth]{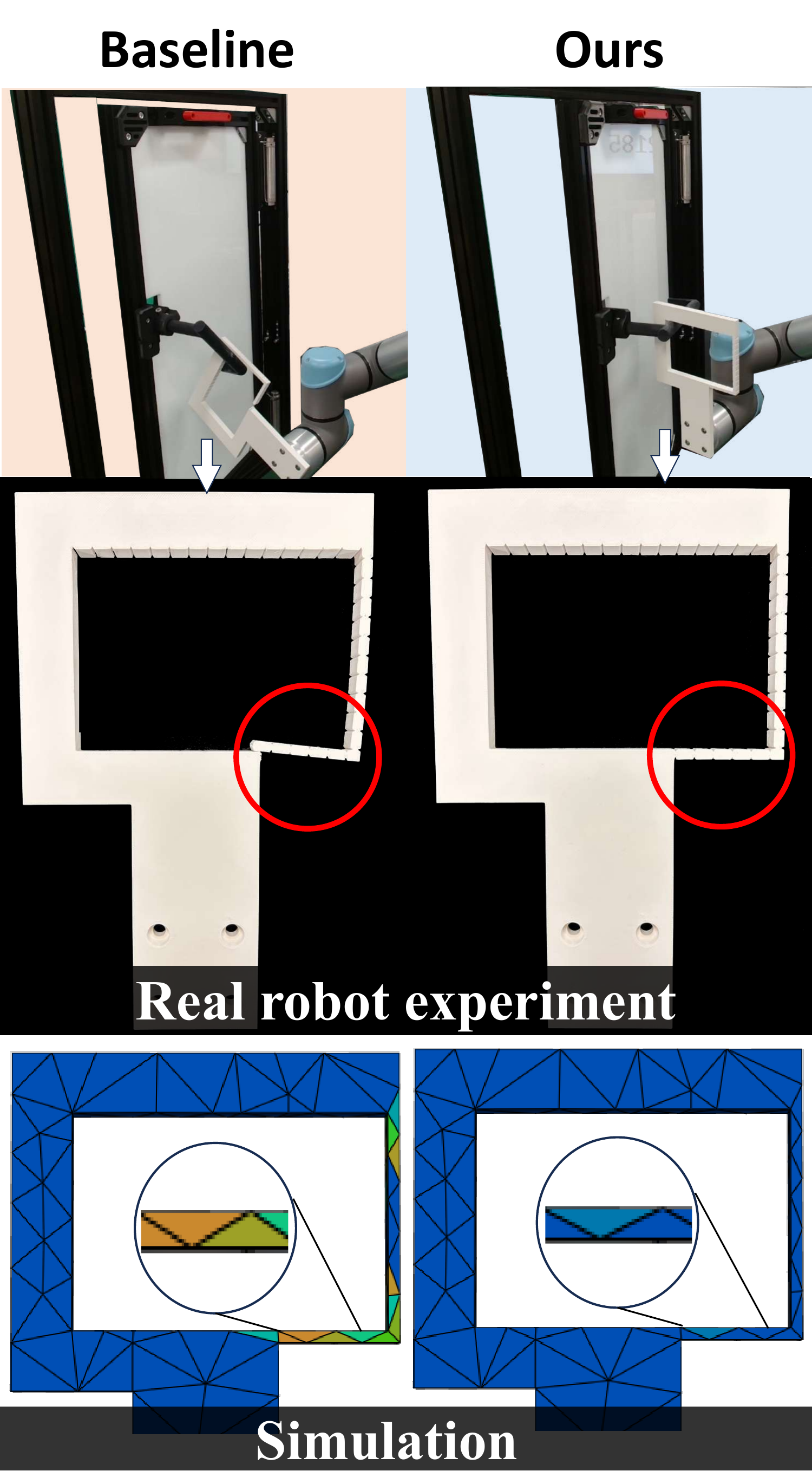}
        \small (a) Tool1
        \label{fig:OD_BT_T1}
    \end{minipage}
    \hfill
    \begin{minipage}{0.49\linewidth}
        \centering
        \includegraphics[width=\linewidth]{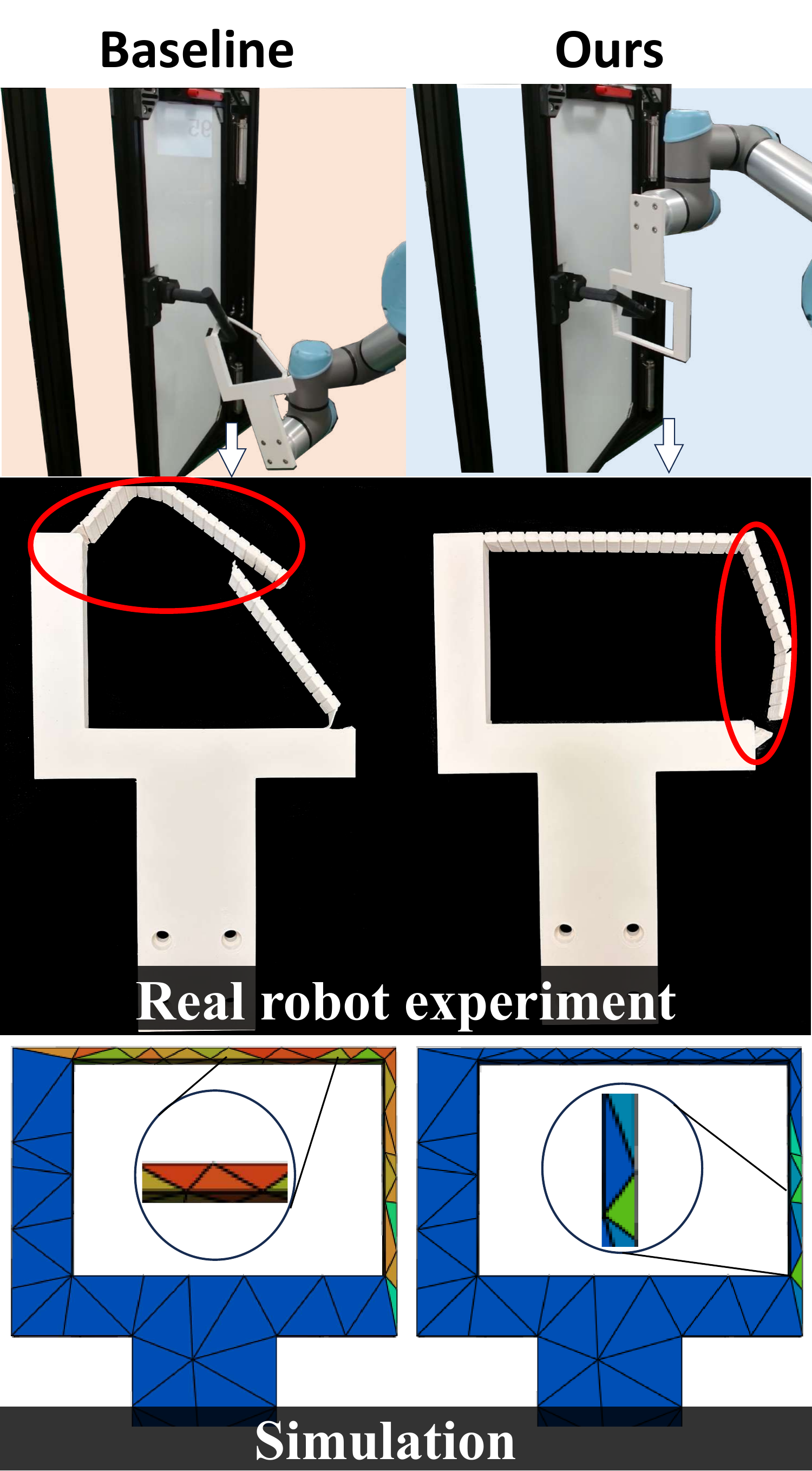}
        \small (b) Tool2
        \label{fig:OD_BT_T2}
    \end{minipage}
    \caption{
    \textbf{Broken tools observed in the Door-Opening experiment with corresponding simulation analysis.} The top row shows results from real-world experiments, highlighting fracture locations after repeated door opening cycles, marked with red circles. The bottom row presents RUL heatmaps from simulation, indicating regions with minimum predicted lifespan. Comparison demonstrates that observed fractures in physical experiments align with the lowest RUL regions identified in simulation analysis.
    }
    \label{fig:Broken_Tool_OD}
\end{figure}

In the Door-Opening task, \emph{\textbf{Ours}} similarly outperformed the baseline, with Tool1 achieving over $3\times$ the execution cycles and Tool2 reaching $1.57\times$ improvement. As shown in \fref{fig:Broken_Tool_OD}, the \emph{\textbf{Baseline}} method for Tool1 resulted in fractures at the handle-rod junction, while \emph{\textbf{Ours}} preserved structural integrity. For Tool2, although both methods exhibited damage, the fracture locations differed: the baseline showed failure at the upper-right rod connection, whereas \emph{\textbf{Ours}} spread usage across stronger regions, resulting in different fracture points consistent with simulated stress predictions (\fref{fig:Door_traj}). 

\subsubsection{Summary}
These results directly address Q5 by demonstrating that the learned policies retain their ability to extend tool lifespan when transferred from simulation to real-world robotic systems. For the Object-Moving task, the observed failure patterns closely align with the minimum RUL regions identified in simulation stress analysis (\fref{fig:Trajetories}), validating the predictive accuracy and practical relevance of our approach. In the Door-Opening task, the lifespan-guided policies learned in simulation effectively transfer to physical executions, consistently extending tool lifespan while maintaining successful task performance. These findings confirm the robustness and real-world applicability of our method.

\begin{table}[t]
    \centering
    \caption{\textbf{Tools' lifespan results in real-world experiments.} This table presents the results from deploying the policies on a physical robot in real-world  scenarios. The values represent the mean number of successful task executions before the tool experienced physical failure, averaged over three independent trials.  For cases of exceptional durability, the experiments were concluded after reaching a predefined maximum number of executions, at which point no failure had been observed. The bottom row shows the improvement factor in execution cycles for our proposed method's policy compared to the Baseline. These real-world results validate that our proposed method significantly prolongs the physical lifespan of the tools.}
    \begin{threeparttable}
        \begin{tabular}{p{1cm} c c c c}
            \toprule
            \multirow{3}{*}{\textbf{Method}} & \multicolumn{2}{c}{\textbf{Object-Moving}} & \multicolumn{2}{c}{\textbf{Door-Opening}} \\
            \cmidrule(lr){2-3} \cmidrule(lr){4-5}
             & {Tool1} & {Tool2} & {Tool1} & {Tool2}\\
            \midrule
            \emph{\textbf{Ours}} & $\mathbf{1682 \pm 88}$ & {\small{over}} $\mathbf{900}$ &  {\small{over}} $\mathbf{1900}$ & $\mathbf{824 \pm 86}$ \\
            \emph{\textbf{Baseline}} & $844\pm 46$ & $281\pm 48$ & $611\pm 11$ & $522\pm 18$ \\
            \midrule
             Improv. & 1.99$\times$ & \small{over} 3$\times$ & \small{over} 3$\times$ & 1.57$\times$\\
            \bottomrule
        \end{tabular}
    \end{threeparttable}
    \label{table:Real_result}
\end{table}

\section{Discussion}\label{sec:discuss}

The proposed lifespan-guided reinforcement learning framework successfully extends the lifespan of the tool across various tasks and tool geometries while maintaining task performance. Despite these promising results, several technical limitations remain, which highlight directions for future improvement.

\subsection{Simulation-Dependence of Lifespan Estimation}

Since FEA is a simulation-based numerical method, our RUL estimation process relies entirely on simulated stress histories. In real-world applications, this creates a practical limitation, as we must either replicate simulation-based stress evaluation in parallel or depend on sim-to-real transfer of pre-trained policies. Without access to real-time structural stress data, the use of RUL feedback in live policy updates becomes infeasible.

\subsection{Lack of Policy Flexibility for Damage-Consideration Strategy Selection}

While the proposed method effectively learns task-completing policies that extend tool lifespan, it currently assumes that a single, fixed policy is sufficient for each task and tool combination. This approach lacks the flexibility to adapt when different tool use strategies may be preferable under varying damage conditions. Although a single optimal usage pattern may indeed offer the best lifespan in some cases, this assumption may not hold in long-term or dynamic usage scenarios, where structural degradation may shift the optimal policy over time.

Moreover, because RUL estimation is only feasible through simulation-based stress analysis, the current framework lacks a mechanism to evaluate or switch between policies based on the tool’s actual condition during deployment. To overcome this limitation, future work could explore incorporating a damage-consideration decision layer, such as hierarchical reinforcement learning~\cite{pateria2021hierarchical} a high-level controller that selects among pre-trained sub-policies depending on estimated wear patterns. This would enable the system to adjust its strategy to maintain efficiency as the damage state evolves.

\subsection{Limited Scalability Across Tasks and Reward Structures}

The proposed method requires training a separate policy for each task, which increases computational cost and hinders scalability. Furthermore, the current design of the ARN mechanism assumes a single reward distribution and does not support multiple concurrent reward structures. This prevents direct generalization to multi-task settings or shared-tool use across diverse applications.

A potential solution is to adopt policy distillation~\cite{Rusu2016distillation}, where multiple task-specific teacher policies are first trained independently, and then compressed into a single student policy. This approach allows the model to retain task-specific expertise while supporting generalization across tasks. When combined with an extended ARN mechanism that preserves per-task reward normalization, such an architecture could facilitate tool use strategies that are both efficient and lifespan-guided across varied scenarios.

\section{CONCLUSION}\label{sec:conclusion}
This paper presents a lifespan-guided tool use reinforcement learning framework that explicitly incorporates the lifespan of a general-purpose tool as a performance criterion alongside task completion. Grounded in the perspective that robotic tool use involves learning task-completing policies under physical restrictions, our method integrates Finite Element Analysis (FEA) and Miner’s Rule to estimate structural damage and guides policy learning toward strategies that balance effectiveness with durability.

Experimental results on Object-Moving and Door-Opening tasks show that our method consistently prolongs general-purpose tool lifespan while maintaining task success, outperforming task-only and torque-minimization baselines. The adaptive reward normalization mechanism enables stable learning without predefined lifespan bounds, and the learned policies successfully transfer from simulation to real-world robots. Moreover, observed real-world tool failures correspond with FEA-predicted weak points, validating both the accuracy and practical utility of the proposed framework. Our method provides an approach for optimizing tool use in scenarios where minimizing wear and extending operational lifespan are essential to sustained robotic performance.

\end{document}